\documentclass[10pt,journal,compsoc]{IEEEtran}
\ifCLASSOPTIONcompsoc
  \usepackage[nocompress]{cite}
\else
  \usepackage{cite}
\fi
\ifCLASSINFOpdf
\else
\fi

\usepackage{stfloats}
\usepackage{graphicx}
\usepackage{amsmath}
\usepackage{amssymb}
\usepackage{booktabs}

\usepackage{multirow}
\usepackage{mathtools}
\usepackage{xcolor}
\usepackage{colortbl}
\usepackage{lipsum}  

\usepackage[capitalize]{cleveref}
\crefname{section}{Sec.}{Secs.}
\Crefname{section}{Section}{Sections}
\Crefname{table}{Table}{Tables}
\crefname{table}{Tab.}{Tabs.}

\renewcommand{\epsilon}{\varepsilon}
\renewcommand{\phi}{\varphi}

\DeclareMathOperator{\argmax}{\arg\,\max}

\newcommand\numberthis{\addtocounter{equation}{1}\tag{\theequation}}

\newcommand{\calD}{\mathcal{D}}
\newcommand{\calL}{\mathcal{L}}

\newcommand{\calA}{\mathcal{A}}

\newcommand{\calP}{\mathcal{P}}
\newcommand{\calI}{\mathcal{I}}
\newcommand{\calN}{\mathcal{N}}
\newcommand{\calF}{\mathcal{F}}
\newcommand{\calH}{\mathcal{H}}

\newcommand{\bbR}{\mathbb{R}}
\newcommand{\bbI}{\mathbb{I}}

\newcommand{\bp}{\mathbf{p}}

\newcommand{\by}{\mathbf{y}}

\newcommand{\bz}{\mathbf{z}}

\newcommand{\bG}{\mathbf{G}}
\newcommand{\bh}{\mathbf{h}}

\newcommand{\feat}{\mathbf{f}}

\newcommand{\vid}{\ensuremath{\text{V}}}
\newcommand{\enc}{\ensuremath{\mathbf{\Phi}}}
\newcommand{\dec}{\ensuremath{\mathbf{\Psi}}}
\newcommand{\bottle}{\ensuremath{\mathbf{\Gamma}}}

\newcommand{\tenc}{\ensuremath{T^{en}}}
\newcommand{\denc}{\ensuremath{d^{en}}}
\newcommand{\fenc}{\ensuremath{\feat^{en}}}

\newcommand{\conf}{\ensuremath{\text{conf}}}
\newcommand{\acc}{\ensuremath{\text{acc}}}

\newcommand{\simi}[2]{\text{cos}\!\left( #1 , #2 \right)}
\newcommand{\EE}{\ensuremath{\mathbb{E}}}

\newcommand{\BK}[1]{ {\left( #1 \right)} }
\newcommand{\sqBK}[1]{ {\left[ #1 \right]} }

\newcommand{\norm}[1]{\left\Vert #1 \right\Vert}
\DeclarePairedDelimiter\ceil{\lceil}{\rceil}
\DeclarePairedDelimiter\floor{\lfloor}{\rfloor}

\newcommand{\losscons}{\ensuremath{\calL_{\text{con}}}}
\newcommand{\losscross}{\ensuremath{\calL_{\text{ce}}}}

\newcommand{\modelname}{C2F-TCN}
\newcommand{\eg}{e.g. }
\newcommand{\ie}{i.e. }

\definecolor{LightCyan}{rgb}{0.88,1,1}
\definecolor{LightPink}{rgb}{1,0.88,1}
\definecolor{LightGreen}{rgb}{0.88,1,0.88}
\definecolor{LightOrange}{rgb}{1,0.94,0.88}

\begin{document}
%
\title{C2F-TCN: A Framework for Semi and Fully Supervised Temporal Action Segmentation}
%
%
%
%

\author{Dipika~Singhania,~\IEEEmembership{Member,~IEEE,}
        Rahul~Rahaman,~\IEEEmembership{Member,~IEEE,}
        and~Angela~Yao,~\IEEEmembership{Member,~IEEE}
\IEEEcompsocitemizethanks{\IEEEcompsocthanksitem Dipika Singhania and Angela Yao is with the School of Computing, National Univeristy of Singapore.\protect\\
E-mail: dipika16@comp.nus.edu.sg, ayao@comp.nus.edu.sg
\IEEEcompsocthanksitem Rahul Rahaman is with the Department of Statistics and Data Science, National University of Singapore. Email: rahul.rahaman@u.nus.edu}
}

%
%

\markboth{Journal of \LaTeX\ Class Files,~Vol.~14, No.~8, August~2015}%
{Shell \MakeLowercase{\textit{et al.}}: Bare Demo of IEEEtran.cls for Computer Society Journals}
%



\IEEEtitleabstractindextext{%
\begin{abstract}
Temporal action segmentation tags action labels for every frame in an input untrimmed video containing multiple actions in a sequence. For the task of temporal action segmentation, we propose an encoder-decoder style architecture named C2F-TCN featuring a ``coarse-to-fine'' ensemble of decoder outputs. The C2F-TCN framework is enhanced with a novel model agnostic temporal feature augmentation strategy formed by the computationally inexpensive strategy of the stochastic max-pooling of segments. It produces more accurate and well-calibrated supervised results on three benchmark action segmentation datasets. We show that the architecture is flexible for both supervised and representation learning. In line with this, we present a novel unsupervised way to learn frame-wise representation from C2F-TCN. Our unsupervised learning approach hinges on the clustering capabilities of the input features and the formation of multi-resolution features from the decoder's implicit structure. Further, we provide first semi-supervised temporal action segmentation results by merging representation learning with conventional supervised learning. Our semi-supervised learning scheme, called ``Iterative-Contrastive-Classify (ICC)'', progressively improves in performance with more labeled data. The ICC semi-supervised learning in C2F-TCN, with 40\% labeled videos, performs similar to fully supervised counterparts. 
\end{abstract}

\begin{IEEEkeywords}
Video analysis, Vision and Scene Understanding, Temporal Action Segmentation, Video Understanding, Temporal Convolution Network, Unsupervised Representation, Semi-Supervised Learning.
\end{IEEEkeywords}}

\maketitle

\IEEEdisplaynontitleabstractindextext

%
\IEEEpeerreviewmaketitle

\IEEEraisesectionheading{\section{Introduction}\label{sec:introduction}}

%
%
%
%



\IEEEPARstart{V}{ideos} of goal-oriented complex activities, \eg~\emph{`frying eggs'}, often have multiple steps or actions in a sequence over time, \eg \emph{`pour oil'}, \emph{`crack egg'}, \ldots, \emph{`put on plate'}.  This work addresses \textbf{temporal action segmentation}, referring to the automatic labeling of each video frame with action labels. Unlike the few-second-long clips used in action recognition~\cite{carreira2017quo}, temporal action segmentation targets longer video sequences that last up to 10 minutes, requiring dedicated architectures.

Feed-forward temporal convolutional networks (TCNs) have proven to be highly effective for temporal action segmentation. 
Two common variants differ in their temporal resolution handling: Encoder-decoders (ED-TCN) ~\cite{TED-lea2017temporal, TEDresi-lei2018temporal, TED-ding2018weakly} shrink and then expand the temporal resolution using layer-wise pooling and upsampling.  Multi-stage architectures (MS-TCN)~\cite{farha2019ms, li2020ms, wang2020boundary, wang2020gated, gao2021global2local} expand the temporal receptive field via dilated convolutions but maintain constant temporal resolutions. ED-TCNs use a single classification stage while MS-TCNs use multiple stages that refine the classification output.

MS-TCNs are now the preferred architecture for action segmentation as they are more accurate than ED-TCNs.  However, their multi-stage architecture is not well-suited to representation learning as the classification refinement process makes it difficult to decouple the representation learning from the classification itself. Class probability refinement stages cannot be learnt with representation learning, and thus the parameters of these stages cannot be trained during this phase.
Therefore, we revisit the ED-TCN architecture to explore its potential for principled feature learning. 

The shrink-and-expand property is particularly noteworthy as it inherently produces multiple temporal resolutions of feature representations. This work leverages multi-resolution to improve the ED-TCN architecture for supervised segmentation and unsupervised representation learning. To this end, we present a new ED-TCN architecture called the Coarse-to-Fine TCN (C2F-TCN),  showing {its generalization capabilities in supervised and semi-supervised temporal action segmentation as well as in complex activity recognition.} \\




\noindent \textbf{C2F-TCN Architecture.}
C2F-TCN, like existing ED-TCNs~\cite{TED-lea2017temporal, TEDresi-lei2018temporal}, follows a U-Net~\cite{ronneberger2015u} style encoder-decoder with 1-D operations in the temporal dimension. At the heart of the C2F-TCN architecture, and its main novelty compared to existing ED-TCNs, is an ensemble of probabilistic decoder outputs. Each successive decoder layer in the ensemble increases in temporal resolution, hence the \textit{``coarse-to-fine''} name. The coarse-to-fine ensemble affords several desirable properties. Firstly, 
it significantly improves segmentation performance by reducing {over-segmentation}, \ie~highly fragmented segmentation outputs. Secondly, it is highly effective at mitigating over-confidence~\cite{rahaman2020uncertainty} and leads to more calibrated segmentation outputs. 

To help train the C2F-TCN, we propose a novel temporal feature augmentation (FA) strategy. Augmentations are common at the image level, with various perturbations and mixes~\cite{zhang2017mixup,yun2019cutmix}. For video, the same augmentations can be applied to each frame ~\cite{videoaug-wang2016temporal, videoaug-han2019video}. 
However, in temporal segmentation, the standard practice is to use snippet-level pre-computed features like IDT~\cite{wang2013action} or I3D~\cite{carreira2017quo} as inputs instead of frames. As such, no previous segmentation works ~\cite{li2020ms, wang2020boundary, wang2020gated, farha2019ms, gao2021global2local} have considered augmentation at feature level for training. This work introduces the first augmentation strategy for temporal action segmentation at the feature level.  
Specifically, 
we sub-sample temporal sequences by stochastically max-pooling segments of input features in time.  
This FA strategy is lightweight, significantly improves the segmentation accuracy across various TCNs, and reduces fragmentation. 
It also leads to better-calibrated segmentation results.  As a result, C2F-TCN with FA surpasses the state-of-the-art for three benchmark segmentation datasets. The proposed framework can also be adapted to recognize complex activity, with accuracies exceeding dedicated models~\cite{highlevel-hussein2020pic,sener2020temporal} by a large margin.\\

\noindent\textbf{Unsupervised Representation Learning.} Equipped with the C2F-TCN architecture and a FA strategy for learning,
we formulate an unsupervised representation learning algorithm suitable for temporal action segmentation.
We are hereby inspired by the success of the contrastive SimCLR framework for images~\cite{simCLR}, videos~\cite{time-contrast-7-qian2020spatiotemporal, time-contrast-3-lorre2020temporal}, and other areas of machine learning~\cite{constrast3-chen2021momentum, contrast6-rahaman2021pretrained}. The standard SimCLR technique brings representations of images~\cite{simCLR} or videos~\cite{time-contrast-7-qian2020spatiotemporal} close to their augmented counterparts during training. However, directly extending this scheme to the long video sequences of temporal action segmentation would incur significant computational expense. Moreover, a direct extension of SimCLR would likely be ineffective, as temporal segmentation models need to capture similarities of semantically similar yet temporally disjoint frames. As the action segments vary in length and content for different video sequences, it is difficult to distinguish whether a feature belongs to the same or different action labels. This makes it non-trivial to incorporate contrastive learning into an unsupervised temporal segmentation task. 

%

In light of this, we design a novel strategy to form the positive and negative sets of contrastive learning without labels. Leveraging the clustering capabilities of the input I3D features enables features in the same cluster to be pulled together while pushing other features apart. An additional advantage of coupling {\modelname} with contrastive learning is the decoder's progressive temporal upsampling. This enables us to form a feature representation that integrates multiple temporal resolutions while enforcing temporal continuity by design. Combining the unsupervised representation learning with supervised segmentation, we formulate a new \emph{semi-supervised setting} that trains on only a small fraction of labeled videos. \\

\begin{figure*}[t!]
    \centering
    \includegraphics[width=0.97\linewidth]{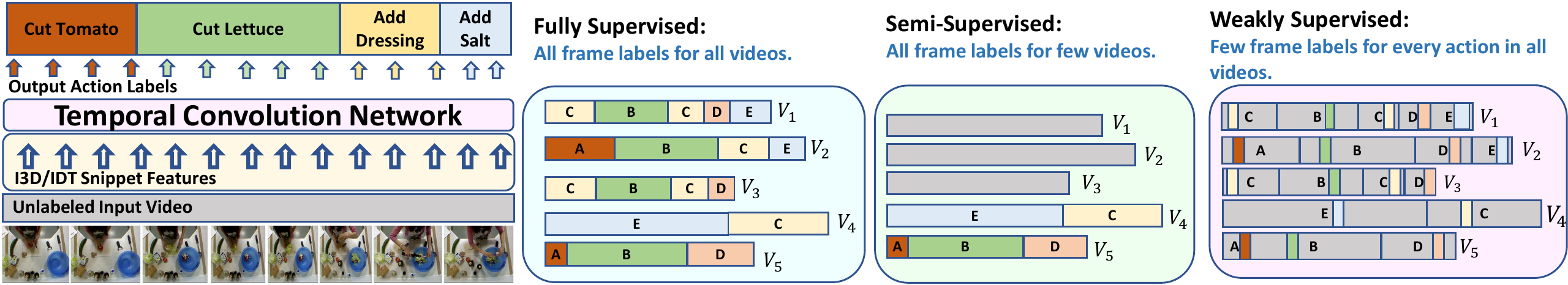}
    \caption{Left: Overview of the temporal action segmentation task with TCN. Right: Comparison of forms of supervision in temporal segmentation.}
    \label{fig:supervision_form}
\end{figure*}

\noindent \textbf{Semi-Supervised Iterative-Contrast-Classify (ICC).}
In temporal action segmentation, a fully supervised setting requires frame-wise labels for every single video.  
To improve the annotation efficiency, we works towards formulating a semi-supervised \textit{''Iterative-Contrast-Classify''} (ICC) that requires labels from only a fraction of the training videos. ICC fully utilizes the labeled and unlabeled data by updating the representations while learning to segment sequences and assigning pseudo-labels to the unlabeled videos. 
We achieve noteworthy segmentation performance with just 5\% labeled videos; with 40\% labeled videos, we almost match full supervision (see ~\cref{fig:teaser_semi}, ~\cref{tab:different_supervsion_breakfast}).  
To the best of our knowledge, our work is the first to apply semi-supervised learning for temporal action segmentation. 
The closest works in spirit~\cite{selfsupervised-chen2020action,timestamp-weakly-li2021temporal} are weakly-supervised and require (weak) labels for \textit{every} training video (see ~\cref{fig:supervision_form} left).  

Our main contributions can be summarized as follows.

\begin{itemize}
\item C2F-TCN, an improved encoder-decoder architecture that features a ``coarse-to-fine'' ensemble of decoder outputs.  C2F-TCN is
flexible for both supervised and representation learning.
\item A model-agnostic temporal FA strategy for segmentation with improvements in accuracy, calibration, and fragmentation. 
\item An unsupervised representation learning approach that leverages clustering and video continuity. The representation learning is enhanced by a novel multi-resolution representation that inherently encodes sequence variations and temporal continuity. 
\item A semi-supervised formulation of temporal action segmentation with an accompanying ICC algorithm that iteratively fine-tunes representations and strengthens segmentation performance with few labeled videos.
\item C2F-TCN, combined with FA, is more calibrated and accurate than the fully supervised state-of-the-art by a large margin. In the semi-supervised setting, our ICC algorithm boasts impressive performance with just 5\% labeled videos; with 40\%, ICC becomes comparable to fully supervised counterparts.



\end{itemize}

Part of this work was first published in~\cite{singhania2022iterative}, where we proposed unsupervised representation learning and its application in the semi-supervised ICC. In this journal extension, we comprehensively show and evaluate the design of the C2F-TCN architecture for fully supervised temporal action segmentation and complex activity recognition. We also introduce a model-agnostic temporal FA strategy that increases the accuracy and calibration of C2F-TCN and other existing TCNs. The C2F-TCN framework is more accurate, is calibrated, and unifies supervised, unsupervised representation and semi-supervised temporal action segmentation. 


\section{Related Work}\label{sec:related_work}

\subsection{Temporal Action Segmentation} 
The task of temporal action segmentation requires information about both 
fine-grained spatio-temporal motion along with long-range temporal patterns in order to parse the action compositions within a complex activity. Local motion information is captured via IDT~\cite{wang2013action} or Kinetics-pretrained I3D~\cite{carreira2017quo} features, which are then further used by segmentation models to capture long-range temporal patterns.
Segmentation models initially consisted of RNNs~\cite{prevsegment-statis-richard2016temporal, prevsegmentrnn-singh2016multi, prevsegmentrnn-perrett2017recurrent, prevsegmentrnn-kuehne2018hybrid}, but they are less effective and slow, especially for long sequences.  Feed-forward TCNs, such as ED-TCN~\cite{TED-lea2017temporal} and its variants~\cite{TED-ding2018weakly,  TEDresi-lei2018temporal} and MS-TCN~\cite{li2020ms, farha2019ms} and its variants~\cite{wang2020boundary, wang2020gated}, perform faster and offer higher performance than RNNs. Throughout this work, we refer to the improved version~\cite{li2020ms} as MS-TCN rather than its earlier version~\cite{farha2019ms}. MS-TCN has been shown to have higher accuracy than  ED-TCN. 
%
This work proposes an improved ED-TCN with novel coarse-to-fine ensembling that is more calibrated \emph{and} accurate.

Following the success of MS-TCN~\cite{li2020ms}, several works have built upon it to improve the over-segmentation, and thus the accuracy, of MS-TCNs. GatedR~\cite{wang2020gated} adds GRU in the refinement stages of MS-TCN. BCN~\cite{wang2020boundary} trains an additional boundary detection model apart from MS-TCN and merges the outputs of the segmentation and boundary model using a post-processing step. ASRF~\cite{ishikawa2021alleviating} extracts features from a fully supervised MS-TCN and builds temporal models and a boundary network over the features extracted from the MS-TCN, thus adding another phase of training and inference after MS-TCN. Unlike these, we handle over-fragmentation with our ensembled prediction in C2F-TCN and the model-agnostic feature-augmentation strategy, thereby eliminating the need for any additional networks or training phases.


\subsection{Varying Supervision in Segmentation} 
The different types of supervision used in temporal action segmentation are illustrated in ~\cref{fig:supervision_form}. 
\textbf{Fully supervised} methods require every-frame annotations for all the videos in the dataset. TCN frameworks include pool-and-upsample style encoder-decoders~\cite{TED-lea2017temporal, TEDresi-lei2018temporal} or 
temporal resolution preserving MS-TCNs~\cite{li2020ms, farha2019ms, wang2020boundary, wang2020gated, gao2021global2local}.  \textbf{Weakly supervised} methods bypass every-frame annotations and use labels such as ordered lists of actions ~\cite{TED-ding2018weakly, weakly-richard2018neuralnetwork, weakly-chang2019d3tw, weakly-li2019weakly, souri2021fast}
or a small percentage of action timestamps (TSS)~\cite{timestamp-weakly-li2021temporal, selfsupervised-chen2020action} for \textit{all} videos. TSS~\cite{timestamp-weakly-li2021temporal} uses single timestamp labels for \textit{every action} in all training videos and SSTDA~\cite{selfsupervised-chen2020action} uses labels for 65\% of timestamps in all training videos. We propose the first \textbf{semi-supervised} setup requiring every-frame annotations, but for only a \textit{few} training videos. Our setup is analogous to semi-supervised image segmentation~\cite{adversarialSSL2DSemanticSeg,ConsistSSL2DSemanticSeg} \ie most training images are un-annotated, while a few are fully annotated. The analogue of TSS~\cite{timestamp-weakly-li2021temporal} is point-supervision~\cite{pointsupervision-bearman2016s}, \ie labeling one pixel from each object of every training image. While TSS requires one frame label for each action and the overall percentage of labeled frames is very small (0.03\%), the annotation effort should not be underestimated. Annotators must still watch \textit{all} the videos, and labeling timestamp frames gives only a 6X speedup compared to densely labeling all frames~\cite{singleframe-sfnet-ma2020sf}. 

Additionally, \textbf{unsupervised} approaches use clustering, including $k$-means~\cite{unsupervised-kukleva2019unsupervised}, agglomerative~\cite{sarfraz2021temporally}, and discriminative clustering~\cite{unsupervised-sener2018unsupervised}. 
To improve clustering performance, some works~\cite{unsupervised-kukleva2019unsupervised, unsupervised-vidalmata2021joint} learn representation by predicting frame-wise features' absolute temporal positions in the video. 
Different from these, in our unsupervised representation learning we implicitly capture the relative temporal relationships based on temporal distance rather than absolute positions.  
Unsupervised clustering approaches can only segment the videos, however, the task of temporal segmentation involves both segmenting and labeling the action segments. Therefore, unsupervised clusters segments evaluated based on Hungarian matching to ground truth labels is not directly comparable to other form of supervision like full supervision or weak supervision. In fact, by using the same representation with two different clustering algorithms, the Hungarian matching segmentation results can vary widely.
Rather than developing an unsupervised clustering algorithm, we develop an \textbf{unsupervised feature learning} task that helps to create discriminative features wherein a simple linear classifier can separate features according to action classes. The linear classifier evaluation protocol is widely used to evaluate unsupervised representation learning tasks~\cite{simCLR, zhang2016colorful, bachman2019learning}, whereby a linear classifier is trained on features from the frozen base network, and test accuracy is used as a proxy to evaluate for learnt representation quality. Beyond linear evaluation, we also evaluate our learning task by fine-tuning for semi-supervised learning; good segmentation results are hereby achieved for training with only a few labeled videos.

\subsection{Complex Activity Recognition} The task of complex activity recognition aims to classify the goal-oriented activity label of the video.
Complex activity recognition~\cite{highlevel-girdhar2017actionvlad, highlevel-hussein2019videograph, highlevel-hussein2019timeception, highlevel-hussein2020pic} follows the strategy of temporal action segmentation (different from strategies of standard short trimmed video recognition~\cite{carreira2017quo, trimmed2d-feichtenhofer2016convolutional, trimmednon-wang2018non, trimmed3d-ji20123d}).  
and uses pre-computed 
snippet-level features as inputs for dedicated sequence-level models designed for recognition~\cite{kuehne2014language, highlevel-girdhar2017actionvlad, highlevel-hussein2019videograph, highlevel-hussein2019timeception, highlevel-hussein2020pic,sener2020temporal}. Given the similarity in the approaches, we posit that the same TCN architecture can be designed for both segmentation and recognition with minimal changes. We here directly use the C2F-TCN's temporal max-pooled final encoder representation
with a linear classifier, which we find to be effective without the need for dedicated architectures.
\begin{figure*}
\begin{center}
\includegraphics[width=\linewidth]{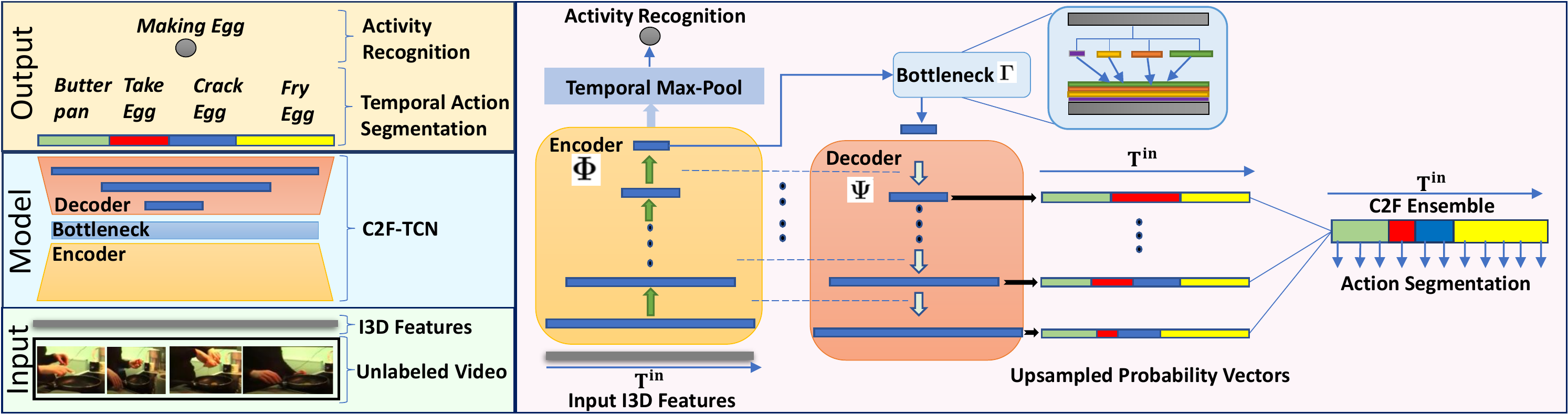}
\end{center}
\caption{Our model \modelname{}, utilizing the decoder's implicit multiple resolution structure to produce \textit{Coarse-to-fine Ensemble} predictions for segmentation; encoder's output for activity recognition.}
\label{fig:main_architecture}
\end{figure*}

\subsection{Unsupervised Contrastive Feature Learning} Contrastive learning dates back to~\cite{contrast6-hadsell2006dimensionality} but was more recently formalized in SimCLR~\cite{simCLR}. Most works~\cite{constrast3-chen2021momentum, contrast6-rahaman2021pretrained, contrast4-he2020momentum, contrast5-khosla2020supervised} hinge on well-defined data augmentations, with the goal of bringing together the original and augmented samples in the feature space. 

The few direct extensions of SimCLR for video~\cite{time-contrast-5-bai2020can, time-contrast-7-qian2020spatiotemporal, time-contrast-3-lorre2020temporal} target action recognition in short clips a few seconds long. Others integrate contrastive learning 
by bringing together next-frame feature predictions with actual representations~\cite{time-contrast-1-kong2020cycle, time-contrast-3-lorre2020temporal}, using path-object tracks for cycle-consistency~\cite{temporal-constrast-wang2020contrastive}, and considering multiple viewpoints~\cite{time-contrast-2-sermanet2018time} or accompanying modalities like audio~\cite{time-contrast-4-alwassel2019self} or text~\cite{time-contrast-6-miech2020end}. These works inspire us to develop contrastive learning for long-range segmentation. However, previous works differ fundamentally in both the aim,~\ie learning the underlying distribution of cycle-consistency in short clips, and input data,~\eg multiple viewpoints or modalities.

\subsection{Multi-Level Resolutions and Scaling} A number of works ranging from semantic segmentation~\cite{FPN-OD-cheng2020panoptic, FPN-OD-lin2017feature} to video temporal action detection~\cite{FPN-TD-li2019deep, sener2020temporal, Multi-Resolution-zhang2019multi} incorporate multiple scales to improve performance. Various ways to incorporate multi-scaling are using multiple sliding windows~\cite{Multi-Resolution-gao2017cascaded, Multi-Resolution-weinzaepfel2015learning}, a multi-resolution attention network~\cite{Multi-Resolution-gao2017cascaded, sener2020temporal}, or multi-resolution loss functions~\cite{FPN-OD-cheng2020panoptic, FPN-OD-lin2017feature, FPN-TD-li2019deep}. Feature Pyramid networks ~\cite{FPN-TD-li2019deep, FPN-OD-lin2017feature} are the most similar to our work in that they use the decoder's implicit layers' multi-resolution feature. However, they utilize multi-resolution outputs from different decoder layers via layer-wise loss functions (choosing the final layer outputs as a prediction). In contrast, our C2F-Ensemble utilizes multi-resolution probabilities by constructing an ensembled prediction probability vector during both training and inference. Our experiments suggest that the proposed ensemble performs better than using loss at every layer. 


\section{Base Segmentation Model \modelname}\label{sec:base_model}

\subsection{Definitions} \label{subsec:definitions} 
We denote a video as $\text{V} \in \bbR^{T \times F}$; for each temporal location $t\!<\!T$, 
frame $\vid[t] \in \bbR^F$ is a $F$-dimensional pre-trained I3D feature.
Note, the input I3D feature is from a model pre-trained on the Kinetics dataset ~\cite{carreira2017quo} and is not fine-tuned on our segmentation datasets. 

Temporal action segmentation (depicted on the right side of \cref{fig:supervision_form}) aims to map each frame feature $\vid[t]$ to an action label $\hat{y}[t] \in \mathcal{A}$, where $\mathcal{A} := \{1,...,C\}$ represents the set of $C$ actions. A temporal segmentation model $M$ takes $\text{V}$ as input and produces predictions $M(\text{V}) = \bp \in \bbR^{T\times C}$, where for each time $t<T$, $\bp[t]\in \bbR^C$ is a probability vector of dimension $C$, and $\bp[t,k]$ denotes the probability assigned to the $k^{th}$ class. The predicted label for each $t$ is then obtained by $\hat{y}[t] = \argmax_k \;\bp[t,k]$ and the corresponding probability by $\hat{p}[t] = \max_k \; \bp[t, k]$ over all possible actions in $k \in \calA$. 
Additionally, for some video datasets (like Breakfast~\cite{kuehne2014language}), each video has a higher-level \textit{complex activity} label $c \in \{1,\ldots,C_{\text{V}}\}$. The complex activity specifies an underlying objective, \eg \emph{`making coffee'} for the action sequence $\{$\emph{`take cup'}, \emph{`pour coffee'}, \emph{`add sugar'}, \emph{`stir'}$\}$.

\subsection{Base TCN}\label{subsec:model_architec}
Our base model is an encoder-decoder TCN with three components, \ie $\mathbf{M} := (\enc:\bottle:\dec)$, with encoder \enc{}, bottleneck \bottle{}, and decoder \dec{} (depicted in \cref{fig:main_architecture}). Compared to previous ED-TCNs architectures~\cite{TED-lea2017temporal, TEDresi-lei2018temporal}, our architecture is deeper, has smaller temporal convolution kernels, and has added skip connections and a bottleneck layer. Our detailed architecture improvement is verified through experimentation, but these changes alone are insufficient to make the ED-TCN architecture a competitive alternative to MS-TCN. The major novel component of our proposed encoder-decoder architecture design (ie. \textit{Coarse-to-Fine(C2F) Ensemble}) is outlined in \cref{subsec:multilayer_ensemble}. 
Our rationale behind improving the ED-TCN architecture for the task of temporal action segmentation is 1) to have an implicit multiple temporal resolution feature structure and 2) to separate the representation learning from the classification layer. Neither of these properties holds for current state-of-the-art MS-TCN~\cite{li2020ms} architecture, as it is a feed-forward TCN with fixed temporal resolution and uses multiple probability refinement stages.

\noindent \textbf{Encoder \enc:} The input to the encoder comprises down-sampled frame-level features ${\text{V}}^{in} \in \bbR^{T^{in}\times F}$. We down-sample (in time) the full video $\text{V}$ of length $T$ to ${\text{V}}^{in}$ of length $T^{in}$ (see ~\cref{subsec:augmentation}). 
The encoder consists of a 1-D convolution unit $\enc^{(0)}$ and six sequential encoder layers $\{\enc^{(u)}\!:\! u\!\le\!6\}$. In the beginning, $\enc^{(0)}$ projects ${\text{V}}^{in}$ to the feature of dimension $T^{in} \times d_0$;
for $u\!\ge\!1$, the outputs of $\enc^{(u)}$ are $\bbR^{T_u \times d_u}$, where $T_u$ and $d_u$ are the temporal and feature dimensions of each layer $u$, respectively.
Each encoder layer has a 1-D temporal convolution and a max-pooling that halves the temporal dimension to $T_u = \ceil{\frac{T^{in}}{2^u}}$. The final encoder output $\fenc$ has a temporal dimension $\tenc := T_6 := \ceil{\frac{T^{in}}{64}}$ and a latent dimension $d_6$.\\ 

\noindent \textbf{Bottleneck \bottle:} 
To ensure flexibility in varying the lengths of input videos and to facilitate temporal augmentation, 
we introduce a temporal pyramid pooling. Pyramid pooling has been used in image recognition and segmentation~\cite{spp_original, deeplabv3, deeplab, dac-spp-unet, spatial-pp} as well as video recognition~\cite{temporal-pp, zheng2019spatial}.
The input to the bottleneck \bottle{} is the final encoder $\enc^{(6)}$'s output $\fenc$ (shown in  ~\cref{fig:main_architecture}). We apply four parallel temporal max-poolings of varying kernel sizes $\{w^{\gamma}_i: i\le 4\}$, reducing $\fenc$'s length to $\floor*{\frac{\tenc}{w^{\gamma}_i}}$. Each feature is then collapsed to a single latent dimension by a shared 1D convolution of kernel size 1 (keeping the temporal dimension fixed) before upsampling back to the original temporal dimension \tenc{}.  Along with $\fenc$, the four features of dimension $\tenc{} \times 1$ are concatenated 
along latent dimension 
to produce a bottleneck output of size ${\tenc{} \times (4+\denc)}$.

\noindent \textbf{Decoder \dec:} The decoder is structurally symmetric to the encoder; it has six layers $\{\dec^{(u)}: u\!\le\!6\}$, each containing an up-sampling unit and a convolution block.  
For each $u$, the up-sampling unit linearly interpolates
inputs to an output of twice the temporal length before concatenating with encoder $\enc^{(6-u)}$'s output via a skip connection. 
The output of the $u^{th}$ decoder block $\dec^{(u)}$, has the temporal dimension $T_{6-u} = \ceil{\frac{T^{in}}{2^{6-u}}}$ and a latent dimension of $128$. The skip connections merge global information from the decoder with local information from the encoder. The final layer $\dec^{(6)}$, with a skip connection from $\enc^{(0)}$, generates an output of size ${T^{in}\times 128}$. During inference, the action predictions $\{\hat{y}_t: t \le T_{in}\}$ are up-sampled back to the full length $T$ to compare with the original ground truth $\textbf{y}=\{y[t]\}^T_{t=1}$ for proper evaluation.

\begin{figure*}
\begin{center}
\includegraphics[width=0.90\linewidth]{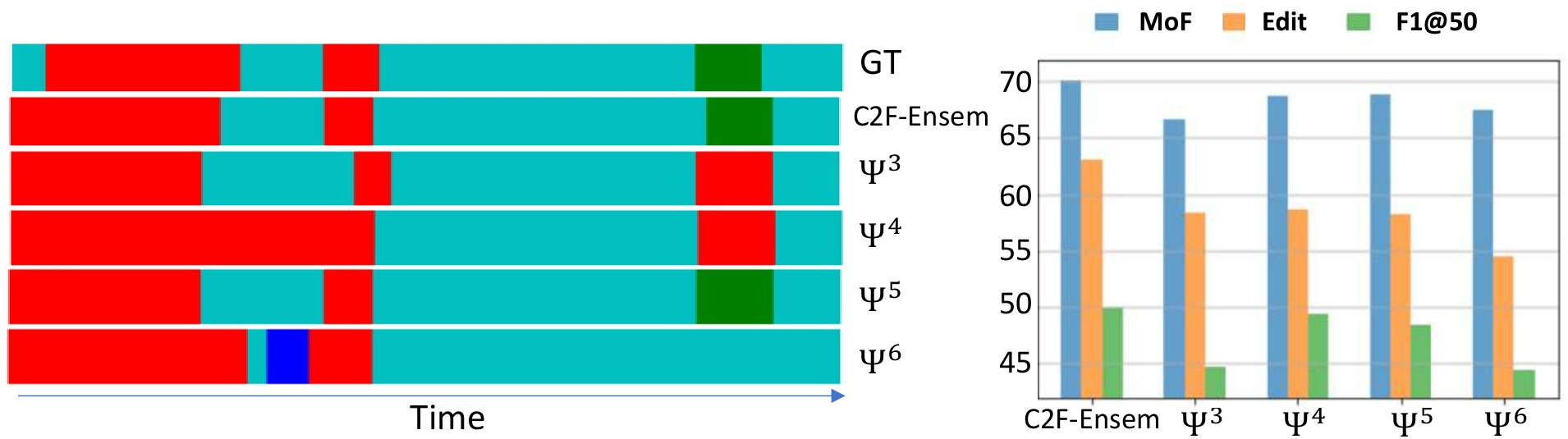}
\end{center}
\caption{Performance of different decoder layers: The left plot shows a qualitative example of our model's segmentation result, where each color denotes an action. We see that C2F-Ensemble (C2F-Ensem) best matches the ground truth (GT) compred to the other layers. Additionally, in C2F-Ensem, the over-fragmentation (blue) patch from the last decoder layer ($\Psi^6$) is removed. The right bar chart shows the quantitative overall performance of the different layers and the C2F-Ensemble. C2F-Ensem has highest Edit, MoF and F1@50.}
\label{fig:layer-analysis}
\end{figure*}

\subsection{Coarse-to-Fine (C2F) Ensemble}\label{subsec:multilayer_ensemble}

Standard encoder-decoders projects the last decoder layer's representation to obtain class probability outputs. We propose ensembling the probability results from several decoder layers. 
We project the representation from the $u^{th}$ decoder block $\dec^{(u)}$ to $C$ dimensions, \ie the number of action classes.  This is followed by a softmax to obtain class probabilities $\bp^{(u)}$ and a temporal upsampling via linear interpolation to the input temporal length $T^{in}$.  Finally, for any $1\!\le\!t\!\le T^{in}$, the ensembled prediction ${\bp}^{ens}[t] \in \bbR^{C}$ is
\begin{gather}\label{eqn:ensemble}
    {\bp}^{ens}[t]\!=\!\sum_{u} \alpha_u \cdot \hat{\bp}^{(u)}[t], \qquad \sum_u \alpha_u\!=\!1, \; \alpha_u\!>\!0
\end{gather}
where $\alpha_u$ is the ensemble weight of the $u^{th}$ decoder and $\hat{\bp}^{(u)} := \text{Up}\sqBK{\bp^{(u)}, T^{in}}$ is the $u^{th}$ decoder probability output $\bp^{(u)}$, up-sampled to a temporal dimension of $T^{in}$. The sum is done action-wise and the final predicted label is calculated as
\begin{equation}
 \hat{y}[t] = \underset{k \in \calA}{\argmax} \, {\bp}^{ens}[t,k].
 \end{equation}
 where ${\bp}^{ens}[t,k]$ is the probability assigned to action class $k$.

\noindent We refer to ${\bp}^{ens}$ as the \textit{coarse-to-fine (C2F) ensemble}; applying it to our base TCN model $\mathbf{M} := (\enc:\bottle:\dec)$ results in our \modelname{}. \textit{`Coarse-to-Fine''} refers to the progressive increase in the temporal resolution of the decoder. 
Decoder layers are ensembled via different upsampling rates.
Our rationale for ensembling is twofold. Firstly, the $\bp^{(u)}$ from earlier decoder layers are inherently coarser in their temporal resolution, making them less susceptible to over-segmentation (since the $\bp^{(u)}$ of temporal dimension $\ceil{\frac{T^{in}}{2^{(6-u)}}}$ is upsampled ${2^{(6 - u)}}$ times to obtain ${T^{in}}$ size outputs). Including them in the ensemble makes it an implicit way to mitigate over-fragmentation errors without the need for additional boundary-detection-model or refinement stages.
Secondly, standard network outputs tend to be over-confident in their predictions, and ensembles are an effective way to reduce overconfidence (see \cref{subsec:uncertainty_quantification}).

\subsection{Temporal Feature Augmentation Strategy}\label{subsec:augmentation}
\textbf{Training augmentations.} 
To augment the sequences, we down-sample the pre-trained feature representations $\text{V}$ and their ground truths $\textbf{y}$. A naive downsampling with random sampling would simply decimate in the temporal dimension features. As an alternative, we opt to use an aggregate (max-pooling) operation over time to create various perturbed features. Max-pooling is computationally efficient yet effective in aggregating video segments~\cite{sener2020temporal}, and we represent multiple temporal resolutions by varying the pooling window. At time $t$, for some temporal window $w$, the pooled feature can be defined as 
\begin{equation}\label{eqn:downsample}
    \vid^{w}[t] = \max_{\tau \in \left[wt, wt+w\right)} \vid[\tau],
\end{equation}

\noindent while taking the ground truth action that is the most frequent in the window $\left[wt, wt+w\right)$ as the corresponding label
\begin{equation}
    y^{w}[t] = \underset{k \in \calA}{\argmax} \sum_{\tau=wt}^{wt+w} \bbI\sqBK{y[\tau]=k},
\end{equation}

\noindent where $\bbI\sqBK{\cdot}$ is the indicator function. The pooled features $\vid^{w}$ and ground truth $\by^{w}$ are of temporal length $T^w\!:=\!\ceil*{\frac{T}{w}}$ and are used as input features ($\vid^{in}$) and ground truth, respectively, during training. 
The augmentation is made stochastic by drawing $w$ from a probability distribution $\pi$.  $\pi$ is parameterized by a base window $w_0$, sampled with a probability of $\pi_0=0.5$, and a uniform distribution over $w$ within a range of $\floor*{\frac{w_0}{2}}$ to $2w_0$:

\begin{gather*}
    \pi = \left\{\begin{array}{ll}
         \pi_0 & : \,w=w_0 \\
         (1 - \pi_0) / (2w_0 - \floor*{\frac{w_0}{2}}) & : \,\floor*{\frac{w_0}{2}}\!\leq\!w\!\leq 2w_0, w\!\neq\!w_0 \\
         0 & : \,\, \text{otherwise} \\ %
    \end{array}\right.
\end{gather*}

\noindent Two advantages of the training augmentation strategy are that it (1) encourages model robustness with respect to a wide range of temporal resolutions and (2) reduces the cost of processing a video down-sampled by window-size $w$ by a factor of $w$.

\noindent \textbf{Test-time augmentations (TTA).} We further leverage augmentations during inference by augmenting the test input features with various $w$ and then combining the predictions after interpolating back to the original temporal length $T$. 
The final predictive probability $\bp^{TTA}$ is estimated as the expected prediction over $\pi$:
\begin{gather}\label{eqn:augmented_predict}
    \bp^{TTA}[t,k] = \EE_{w\sim \pi} \Big[\bp^{ens}[t,k\,|\vid^{w}] \Big],
\end{gather}
where $\bp^{ens}[t,k\,|\vid^{w}]$ is the ensemble probability computed with the input feature $\vid^{w}$.

\subsection{Calibration}\label{subsec:calibration_notation}

Calibration measures the over/under-confidence of predictions. A given prediction $\hat{y}[t] = \argmax_k \bp[t,k]$ has a \textbf{\textit{confidence}} $\hat{p}[t] := \max_k \bp[t, k]$,\ie{} the maximum probability prediction. The associated \textbf{\textit{accuracy}} of a confidence, $\acc(p)$, is the action classification accuracy for all frames with $\hat{p}[t] = p$. Ideally, \acc{} should be high for high confidences and vice versa. A model is calibrated if $\acc(p)\!=\!p, \,\forall p\in [0,1]$; it is over-confident (or under-confident) if $\acc(p)\! \le\!p$ (or $\acc(p)\!>\!p$). 
The \acc{} for a range of confidence $\calP \subset [0,1]$ is defined as
\begin{gather*}
    \acc(\calP) := \frac{\sum_t \bbI\Big[ \hat{y}[t] = y[t] \Big] \cdot \bbI\Big[ \hat{p}[t] \in \calP \Big]}{\sum_t \bbI\Big[ \hat{p}[t] \in \calP \Big]}.
\end{gather*}

The confidence $\calP$ denoted as $\conf(\calP)$ is the average of all the confidence values within $\calP$. We use these notions to later measure the calibration performance in ~\cref{subsec:uncertainty_quantification}.

Calibration is a neglected aspect of temporal action segmentation, and our work is the first to point this out
as well as provide remedies for it.
We show that the standard models are extremely over-confident(~\cref{fig:uncertainty}). Our two main contributions (coarse-to-fine ensemble and the temporal FA strategy) not only result in improved segmentation performance but also significantly improve the calibration (~\cref{fig:uncertainty}).

\section{Fully Supervised Framework}\label{sec:losses}

\subsection{Temporal Action Segmentation}\label{sec:TAS} We use a standard frame-level cross-entropy loss $\calL_{\text{CE}}$ and transition loss $\mathcal{L}_{\text{TR}}$, as per previous works~\cite{li2020ms, wang2020boundary}:
\begin{equation}\label{eqn:cross_entropy}
    \mathcal{L}_{\text{CE}} = - \frac{1}{T} \sum_t \sum_{k\in \calA} \bbI\sqBK{y[t] = k} \cdot \log \bp^{ens}[t,k],
\end{equation}
where $y[t]$ is the ground truth label and $\bp^{ens}[t,k]$ is the \textit{coarse-to-fine} probability (see \cref{eqn:ensemble}) for class $k$. The transition loss $\mathcal{L}_{\text{TR}}$ encourages the same action label in neighbouring frames:
\begin{equation}\label{eqn:transfer_loss}
    \mathcal{L}_{\text{TR}}=\frac{1}{T} \sum_{t} \sum_{k} \min\BK{\delta[t, k], \epsilon_{max}}^2, 
\end{equation}
where, $\delta[t, k] := \left|\log \bp^{ens}[t,k] - \log \bp^{ens}[t-1,k]\right|$ is the inter-frame difference of log-probabilities, and $\epsilon_{max} > 0$ is the maximum threshold for $\delta[t, k]$. We use a joint loss $\calL = \mathcal{L}_{\text{CE}} + \lambda_{\text{TR}} \mathcal{L}_{\text{TR}}$ with $\lambda_{\text{TR}}\!=\!0.15$, $\epsilon_{max}=4$ as used in~\cite{li2020ms}. 

We apply a single loss directly to the ensembled output; 
While other frameworks, like MS-TCN and FPN~\cite{FPN-OD-cheng2020panoptic, FPN-OD-lin2017feature, FPN-TD-li2019deep}, apply loss to each stage individually and use only final layer predictions at inference. The equivalent of applying losses to the up-sampled outputs of each decoder layer has less performance benefit than the proposed C2F-Ensemble (shown in \cref{tab:ensemble_vs_loss_every}). 

\subsection{Complex Activity Recognition} 
We adapt \modelname{} by temporally max-pooling the final encoder representation $\feat^{en}$ (from $\enc^{(6)}$ of temporal dimension $T_6 = \ceil{T^{in}/64}$) over time, obtaining a video-level representation. After that, we apply a two-layer MLP followed by a softmax $\sigma$ to obtain 
\begin{equation}
     \bp_{\text{V}} = \sigma \sqBK{\text{MLP}\sqBK{\max_t (\feat^{en}[t])}},
\end{equation}

\noindent where $\bp_{\text{V}} \in \bbR^{C_{\text{V}}}_{+}$ is the probability vector for the $C_{\text{V}}$ complex activities. Intuitively, max-pooling retains the important information without needing to consider the order of actions in the complex activity. Similar to~\cite{highlevel-hussein2019timeception, highlevel-hussein2020pic}, 
we train this network without frame-wise action cross-entropy loss and apply only the following video-level cross-entropy loss:
\begin{equation}
 \mathcal{L}_{\vid} = - \sum_{c\in \calA_{\text{V}}} \bbI\sqBK{y_{\text{V}} = c} \cdot \log \bp_{\text{V}} \sqBK{c}.
\end{equation}

\noindent where $y_{\text{V}}$ is the ground truth, and  $\bp_{\text{V}} \sqBK{c}$ is the predicted probability assigned to complex activity class $c$ for video $\text{V}$.

\section{Semi-Supervised Framework} \label{sec:preliminaries} In the semi-supervised framework (illustrated in~\cref{fig:supervision_form}) we use only a small fraction of labelled training videos instead of using all labelled training video as described in fully supervised setting (~\cref{sec:TAS}). 
We formulate the semi-supervised learning using our base temporal segmentation model 
the C2F-TCN (see ~\cref{sec:base_model}), although our method is also applicable to other base encoder-decoder models such as ED-TCN~\cite{TED-lea2017temporal}.
For simplicity, unless otherwise explicitly noted, \eg in ~\cref{subsec:multi-resolution-similarity}, we treat the temporal dimension of all the videos as a normalized unit interval $t \in [0,1]$,~\ie $T\!=\!1$. As before, each video frame $\vid[t]$ has a ground truth action label $y[t] \in \calA := \{1,...,C\}$ from a pre-defined set of $C$ action classes.  

\begin{figure*}[t!]
\centering
  \includegraphics[width=0.90\textwidth]{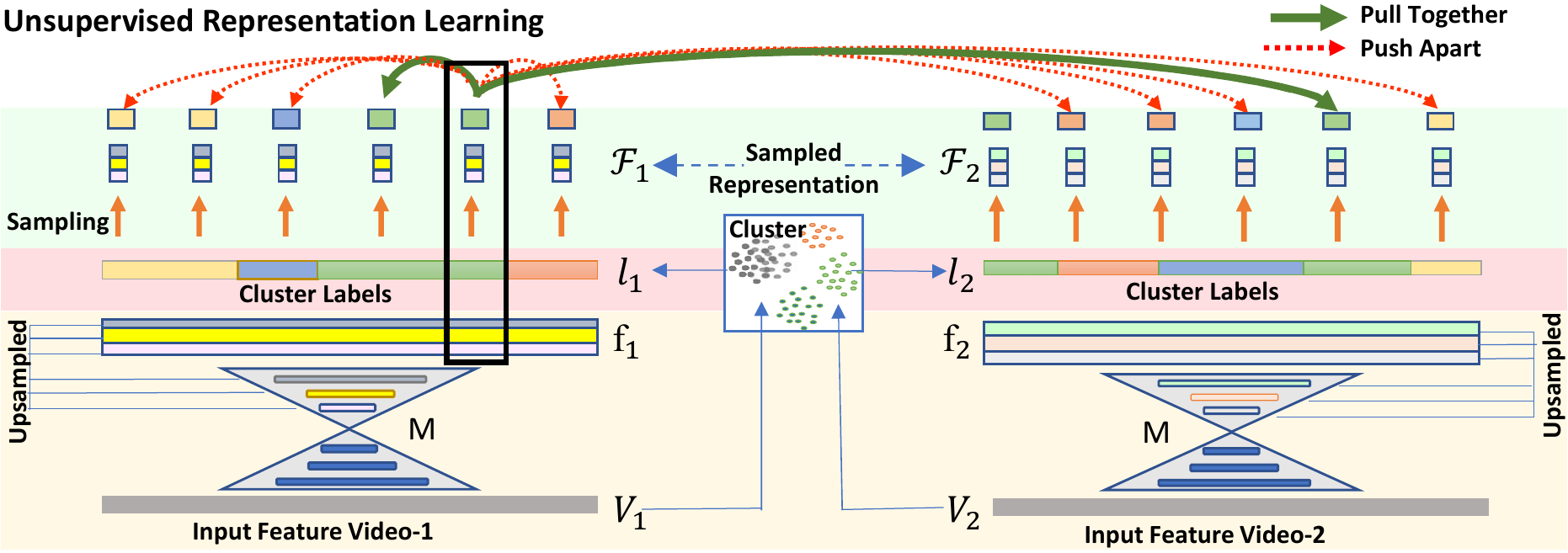}
  \caption{Unsupervised Representation Learning Depiction: Step 1 (bottom orange panel): Pass pre-trained I3D inputs $V$ into the base TCN and generate a multi-resolution representation $\feat$. Step 2 (middle pink panel): Cluster the I3D inputs $V$ within a training mini-batch and generates frame-wise cluster labels $l$. Step 3 (top green panel): Representation $\feat$ and its corresponding cluster label $l$ are sampled based on a temporal proximity sampling strategy to form feature set $\mathcal{F}$. Step 4: Apply contrastive learning to ``pull together'' (green arrows) similar samples in the positive set and ``push apart'' (red arrows) other samples in the negative set.} 
 \label{fig:contrastive_pic}
\end{figure*}


\textbf{Learning Framework \& Data Split:}\label{subsec:framework_splits}
Our semi-supervised framework has two stages.  First, we apply an unsupervised representation learning to learn model $\mathbf{M}$ (~\cref{subsec:unsupervised_representation_learning}). Subsequently, model $\mathbf{M}$ is trained (fine-tuned) with linear projection layers (action classifiers) on a small portion of the labeled training videos to produce the semi-supervised model ($\mathbf{M}:\mathbf{G}$)
(~\cref{sec:semi-supervised-temporal-segmentation}). For representation learning, we follow the convention of previous unsupervised works~\cite{unsupervised-kukleva2019unsupervised, unsupervised-vidalmata2021joint} 
in which actions $y$ are unknown but the complex activity of each video $y_{\text{V}}$ \emph{is} known\footnote{The label is used implicitly, as the unsupervised methods are applied to videos of each complex activity individually.}.  For the semi-supervised stage, the ground truth $y$ is used for a small subset of labeled video $\calD_L$ out of a larger training dataset $\calD = \calD_U \cup \calD_L$, where $\calD_U$ denotes the unlabeled videos. 

\textbf{Contrastive Learning}
We use contrastive learning for our unsupervised frame-wise representation learning. Following the formalism of~\cite{simCLR}, we define a set of features $\calF\!:=\!\{\feat_i, i\!\in\!\calI\}$ indexed by a set $\calI$. Each feature $\feat_i\!\in\!\calF$ is associated with two disjoint sets of indices $\calP_i\!\subset\!\calI\!\setminus\!\{i\} $ and $\calN_i\!\subset\!\calI\!\setminus\!\{i\}$. The features in the positive set $\calP_i$ should be similar to $\feat_i$, while the features in the negative set  $\calN_i$ should be contrasted with $\feat_i$. For each $j \in \calP_i$, the contrastive probability $p_{ij}$ is defined as

\begin{equation}\label{eqn:contrastive_prob}
    p_{ij} = \frac{e_{\tau}\BK{\feat_i, \feat_j}}{e_{\tau}\BK{\feat_i, \feat_j} + \sum_{k \in \calN_i} e_{\tau}\BK{\feat_i, \feat_k}},
\end{equation}

\noindent where the term $e_{\tau}=\exp\{\text{cos}(\feat_i,\feat_j)/\tau\} $ is the exponential of the cosine similarity between $\feat_i$ and $\feat_j$ scaled by temperature $\tau$. Maximizing the probability in Eq.~\eqref{eqn:contrastive_prob} ensures that $\feat_i, \feat_j$ are similar while also decreasing the cosine similarity between $\feat_i$ and any feature in the negative set.  The key to effective contrastive learning is to identify the relevant positive and negative sets to perform the targeted task. 

\subsection{Unsupervised Representation Learning}\label{subsec:unsupervised_representation_learning}
We apply contrastive learning at the frame level, based on input feature clustering and temporal continuity (~\cref{sec:frame_level_contratsive}), and at the video-level, by leveraging the complex activity labels (~\cref{subsec:video-level-contrast}).  The two objectives are merged into a common loss that is applied to our multi-temporal resolution feature representations (~\cref{subsec:multi-resolution-similarity}).

\subsubsection{Frame-Level Contrastive Formulation}\label{sec:frame_level_contratsive}

\textbf{Input Clustering:}\label{subsubsec:input_feature_cluster}
Our construction of positive and negative sets should respect the distinction between different action classes.  But, as the setting is unsupervised, there are no labels to guide the formation of these sets. Hence, we propose leveraging the discriminative properties of the pre-trained input I3D features to initialize the positive and negative sets.  The clusters are formed on the input features, but the contrastive learning is done over the representation $\feat$ produced by the C2F-TCN model (yellow panel in~\cref{fig:contrastive_pic}). 

Specifically, we cluster the individual frame-wise inputs $\vid[t]$ for all the videos within a small batch. We use k-means clustering and set the number of clusters as $2C$ (ablations in Apepndeix-C), \ie{} twice the number of actions, to allow variability even within the same action. After clustering, each frame $t$ is assigned the cluster label $l[t] \in \{1,\ldots,2C\}$.  
Note that this simple clustering does not require videos of the same (or different) complex activities to appear in a mini-batch.  It also does not incorporate temporal information -- this differs from previous unsupervised works~\cite{unsupervised-kukleva2019unsupervised, unsupervised-vidalmata2021joint} that embed absolute temporal locations into the input features \textit{before} clustering. 

\textbf{Representation Sampling Strategy:} \label{sec:representation_sampling}  
The videos used for action segmentation are long, \ie 1-18k frames. Contrasting all the frames of every video in a batch would be too computationally expensive to consider, whereas contrastive loss of even a few representations back-propagates through the entire hierarchical TCN. To this end, a fixed number of frames are dynamically sampled from each video to form the feature (representation) set $\calF$ for each batch of videos (shown in green panel of~\cref{fig:contrastive_pic}). Note that the sampling is applied to the feature representations $\feat=\mathbf{M}(\vid)$, and not to the inputs $\vid$, and that the full input $\vid$ is required to pass through the TCN to generate $\feat$. 

Let $\calI$ denote the feature set index (as in sec \ref{sec:preliminaries}), and for any feature index $(n,i) \!\in\! \calI$, let $n$ denote the video-id and $i$ the sample-id within that video. For a video $\vid_n$ and a fixed $K\!>\!0$, we sample $2K$ frames $\{t^n_i: i\!\le\! 2K\} \subset [0,1]$ and obtain the feature set $\calF_n\!:=\!\{\feat_n[t^n_i]\!:\!i\!\le\!2K\}$.  To do so, we divide the unit interval $[0,1]$ into $K$ equal partitions 
and randomly choose a single frame from each partition. Another $K$ frames are then randomly chosen 
$\epsilon$ away ($\epsilon \ll 1/K$)
from each of the first $K$ samples. This strategy ensures diversity 
(the first $K$ samples) while having nearby $\epsilon-$distanced features (the second $K$ samples). This aim is to either enforce temporal continuity, if they are the same action, or learn boundaries, if they are different actions (approximated by the cluster labels $l$ when actions labels are unknown). 

\textbf{\textbf{Frame-Level Positive and Negative Sets:}}\label{subsubsec:positive-negative-sets} Constructing the positive and negative set for each index $(n,i) \in \calI$ requires a notion of similar features. The complex activity label is a strong cue, as there are either few or no shared actions across the different complex activities.  For video $\vid_n$ with complex activity $c_n$, we contrast index $(m, j)$ with $(n, i)$ if $c_m \neq c_n$. In datasets without meaningful complex activities (50Salads, GTEA), this condition is not applicable. 

The cluster labels $l$ of the input features 
already provides some separation between actions (see Table \ref{tab:ablation_unsupervised_representation}); we impose an additional temporal proximity condition to minimize the possibility of a different action in the same cluster. Formally, we bring the representation with index $(n,i)$ close to $(m,j)$ if their cluster labels are the same, i.e $l_n[t^n_i] = l_m[t^m_j]$, \emph{and} if they are close-by in time, ~\ie,~$|t^n_i - t^m_j| \leq \delta$. For datasets with significant variations in the action sequence, \eg 50Salads, the same action may occur at very different parts of the video; 
thus, we choose higher $\delta$, vs. smaller $\delta$, for actions that follow more regular ordering, \eg Breakfast.
Sampled features belonging to the same cluster, but exceeding the temporal proximity, \ie $l_n[t^n_i] = l_m[t^m_j]$ but $|t^n_i - t^m_j| > \delta$, are not considered for either the positive or the negative set.

Putting together the criteria from complex activity labels, clustering and temporal proximity, our positive set ($\calP_{n,i}$) and negative set ($\calN_{n,i}$) for index $(n,i)$, are defined as
\begin{align*}
    \calP_{n,i}\!&=\!\{(m, j)\!: c_m=c_n, |t^n_i - t^m_j| < \delta,\, l_n[t^n_i] = l_m[t^m_j]\} \\
    \calN_{n,i} &= \{(m, j)\!: c_m \neq c_n\} \,\cup \numberthis{} \label{eqn:positive-negative}\\
    &\qquad \{(m,j)\!: c_m=c_n, l_n[t^n_i] \neq l_m[t^m_j]\}
\end{align*}
where $m, n$ are video indices, $t^n_i$ is the frame-id corresponding to the $i^{th}$ sample of video $n$, 
$c_n$ is the complex activity of video $n$, and $l_n[t^n_i]$ the cluster label of frame $t^n_i$.  
For an index $(m, j) \in \calP_{n,i}$, \ie{} belonging to the positive set of $(n,i)$, the contrastive probability becomes
\begin{gather}\label{eqn:frame-level-contrast}
    \!\!\!\! p^{nm}_{ij}\!=\!\frac{e_{\tau}\Big(\feat_n[t^n_i], \feat_m[t^m_j]\Big)}{e_{\tau}\Big( \feat_n[t^n_i], \feat_m[t^m_j] \Big) +\!\!\sum\limits_{(r,k) \in \calN_{n,i}}\!\!\!\!  e_{\tau} \Big( \feat_n[t^n_i], \feat_r[t^r_k]\Big)}.
\end{gather}
where $e_{\tau}$ is the $\tau$-scaled exponential cosine similarity of Eq.~\eqref{eqn:contrastive_prob}. For a feature representation $\feat_n[t^n_i]$, ~\cref{fig:contrastive_pic} visualizes the positive set with pull-together green arrows and negative set with push-apart red arrows. 

\subsubsection{{Video-Level Contrastive Formulation}} \label{subsec:video-level-contrast}
To further emphasize global differences between different complex activities, we construct video-level summary features $\bh_n \in \bbR^d$ 
by max-pooling the frame-level features $\feat_n \in \bbR^{T_n\times d}$ along the temporal dimension.  For video $\vid_n$, we define video-level feature $\bh_n = \max_{1 \le t \le T_n} \feat_n[t]$. Intuitively, the max-pooling captures permutation-invariant features and has been found to be effective for aggregating video segments~\cite{sener2020temporal}. 
With features $\bh_n$, a video-level contrastive learning is formulated. Reusing the index set as video-ids, $\calI = \{1,...,|\calD|\}$, we define a feature set $\calH := \{\bh_n\!: n \le |\calD|\}$, where for each video $n$, there is a positive set $\calP_n := \{m\!: c_m = c_n\}$ and a negative set $\calN_n = \calI \setminus \calP_n$. For video $n$ and another video $m \in \calP_n$ in its positive set, the contrastive probability can be defined as
\begin{gather}\label{eqn:video-level-contrast}
    p_{nm} = \frac{e_{\tau}\BK{\bh_n, \bh_m}}{e_{\tau}\BK{\bh_n, \bh_m} + \sum_{r \in \calN_n} e_{\tau}\BK{\bh_n, \bh_r}}.
\end{gather}

For our final \textbf{unsupervised representation learning} we use a \textbf{contrastive loss} function $\losscons$ that sums the video-level and frame-level contrastive losses: 
\begin{gather}\label{eqn:contrastice-loss}
    \!\!\losscons\! =\! - \tfrac{1}{N_1} \sum_n\!\!
    \sum_{m\in\!\calP_n} \log p_{nm}\! -\! \tfrac{1}{N_2}\!\! \sum_{n,i} \sum_{{m,j} \in \calP_{n,i}} \!\!\!\! \log p^{nm}_{ij},
\end{gather}
where $N_1 = \sum_n |\calP_n|, N_2 = \sum_{n,i} |\calP_{n, i}|$, and $p^{nm}_{ij}, p_{nm}$ are as defined in equation \eqref{eqn:frame-level-contrast} and \eqref{eqn:video-level-contrast} respectively. In practice, we compute this loss over mini-batches of videos.

\subsubsection{{Multi-Resolution Representation}}\label{subsec:multi-resolution-similarity}
This work shows that constructing an appropriate representation can significantly boost the performance of contrastive learning.
For this subsection, we switch to an absolute integer temporal index, \ie for a video $\vid$ the frame indices are $t \in \{1,\ldots, T\}$, where $T \ge 1$. Recall that the decoder layer $\dec{}$ of our C2F-TCN has six layers; each layer $\dec^{(u)}$ produces features $\bz_u, 1\!\le\!u\!\le\!6$ while progressively doubling the temporal resolution, \ie the length of $\bz_{u}$ is  $\ceil*{T/2^{6-u}}$. 
The temporally coarser features provide more global sequence-level information, while the temporally fine-grained features contain more local information.  

To leverage the full range of resolutions, we combine $\{\bz_1,\ldots,\bz_6\}$ into a new feature $\feat$.  Specifically, we upsample each decoder feature $\bz_u$ to ${\hat{\bz}}_u := \text{Up}\BK{\bz_u, T}$ having a common length $T$ using a temporal up-sampling function  $\text{Up}\BK{\cdot, T}$, \ie with \textit{`linear'} interpolation. 
The final frame-level representation for frame $t$ is defined as $\feat[t] = \BK{\bar{\bz}_1[t]:\bar{\bz}_2[t]:\ldots:\bar{\bz}_6[t]}$, where $\bar{\bz}_u[t] = {\hat{\bz}}_u[t]/\norm{{\hat{\bz}}_u[t]}$, ~\ie ${\hat{\bz}}_u[t]$ is normalized and then concatenated along the latent dimension for each $t$ (see ~\cref{fig:contrastive_pic}). 
It immediately follows that for frames $1\!\le\! s,t\! \le\! T$, the cosine similarity $\text{cos}(\cdot)$ can be expressed as
\begin{gather}\label{eqn:similarity-linear-comb}
    \text{cos}(\feat[t],\feat[s]) = \sum_{u=1}^6 \omega_u \cdot \text{cos}(\bz_u[t],\bz_u[s]).
\end{gather}
As a result of our construction, 
the weights in Eq. \ref{eqn:similarity-linear-comb} become $\omega_u = \frac{1}{6}$, \ie each decoder layer makes an equal contribution to the cosine similarity. Normalizing after concatenation, 
would cause Eq.~\eqref{eqn:similarity-linear-comb}'s coefficients $\omega_u \propto \norm{\bz_u[t]}\cdot \norm{\bz_u[s]}$. 
The importance of this ordering is verified in Appendix C.

\textbf{Advantages:}\label{subsubsec:advantages} 
Our representation $\feat$ implicitly encodes some degree of temporal continuity by design. In `\textit{nearest}' up-sampling, it can be shown that for frames $1\!\le\! s,t\! \le\! T$, if $\floor*{t/2^u} = \floor*{s/2^u}$ for some integer $u$, it is implied that $\simi{\feat[t]}{\feat[s]} \ge 1 - u/3$ (detailed derivation in Appendix C).
Including temporally coarse features like $\bz_1$ and $\bz_2$ allows the finer-grained local features $\bz_6$ to disagree with nearby frames without harming the temporal continuity.  This makes the representations less prone to the common occurring problem of over-segmentation. This is demonstrated by the significant improvement in the Edit and F1 scores in the last row of~\cref{tab:multi-resolution-improvement}.
\\

\subsubsection{{Evaluating the Learned Representation}}\label{subsec:linear-evaluation}
To evaluate the 
learned representations, we train a linear classifier $\mathbf{G}_f$ on $\feat$ (\ie representations extracted from frozen C2F-TCN unsupervised trained model) to classify frame-wise action labels. This form of evaluation is directly in lines with widely used linear evaluation protocols of unsupervised representation learning~\cite{temporal-contrast-feichtenhofer2021large, simCLR, SWAV}. The assumption is that if the unsupervised learned features are sufficiently strong, then a simple linear classifier is sufficient to separate the action classes, and test accuracy is used as a proxy for evaluating representation quality.
While our representation learning is unsupervised, evaluation classifier $\mathbf{G}_f$ is fully supervised (\ie with cross-entropy loss $\losscross$), using ground truth labels $y$ over the standard datasets' splits. 
\begin{figure}
\centering
\includegraphics[width=0.8\linewidth]{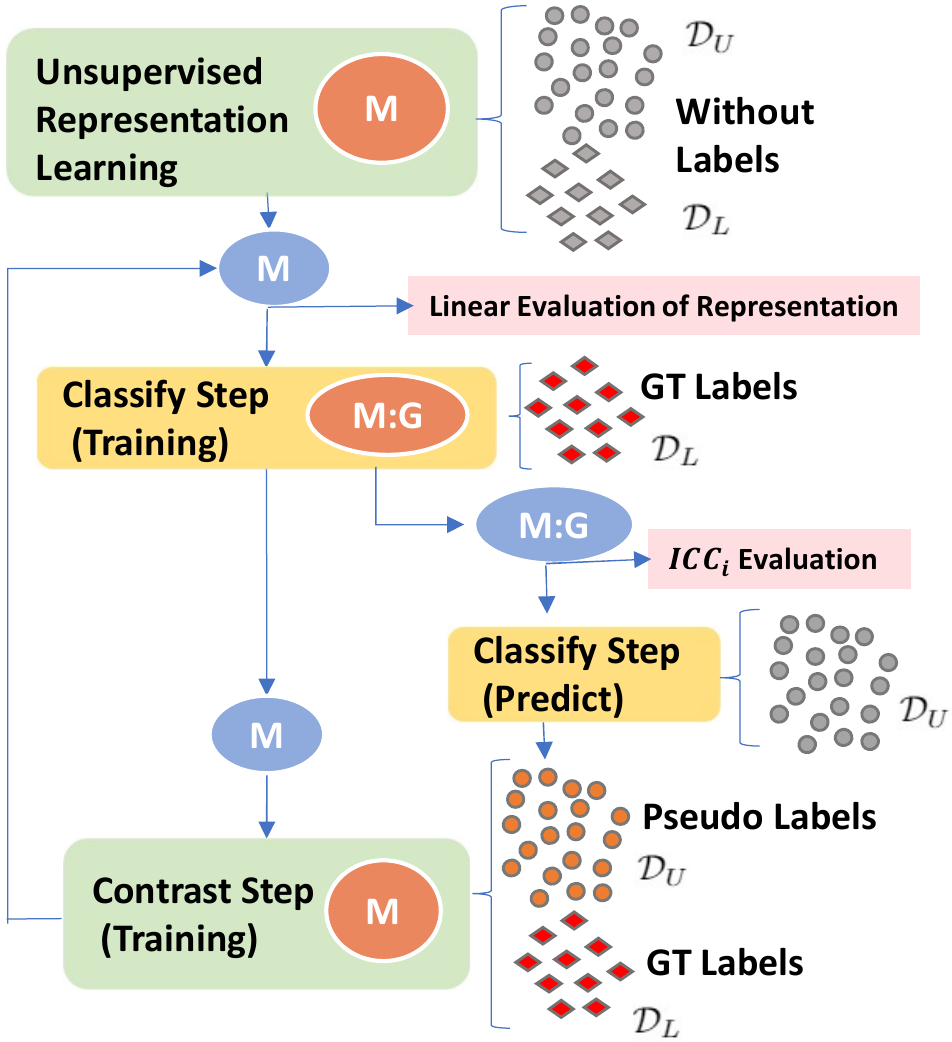}
\caption{Depiction of \textit{Iterative-Contrast-Classify} algorithm}
\label{fig:ICC-depiction}
\end{figure}
\subsection{Semi-Supervised Temporal Segmentation}\label{sec:semi-supervised-temporal-segmentation}
After unsupervised representation learning, model $\mathbf{M}$ cannot yet be applied for action segmentation.  The decoder output must be coupled with a linear projection $\mathbf{G}$ and a softmax to generate the actual segmentation.  $\mathbf{G}$ can only be learned using labels, \ie from $\mathcal{D}_L$, though the labels can be further leveraged to fine-tune $\mathbf{M}$ (~\cref{subsec:classify-step}).  Afterwards, $\mathbf{M}$ and $\mathbf{G}$ can be applied to unlabeled data $\mathcal{D}_U$ to generate pseudo-labels.  The pooled set of labels from $\mathcal{D}_L \cup \mathcal{D}_U$ can then be applied to update 
$\mathbf{M}$ (\cref{subsec:contrast-step}).  By cycling between these updates, we propose an \emph{Iterative-Contrast-Classify (ICC)} algorithm (\cref{subsec:Iterative-Contrast-Classify}) that performs semi-supervised action segmentation (see overview in ~\cref{fig:ICC-depiction}).



\subsubsection{\textit{Classify Step: Learning $\mathbf{G}, \mathbf{M}$ with $\mathcal{D}_L$}}\label{subsec:classify-step}
Similar to the supervised C2F-TCN, 
each 
decoder layer's representations ${\bz}_u$ (temporal dim $\ceil*{T/2^{6-u}}$) is projected with a linear layer $\bG_u$ to $C$-dimensional vector, where $C$ is the number of action classes. This is followed by a softmax to obtain class probabilities $\bp_{u}$ and a linear interpolation in time to up-sample back to the input length $T$. For frame $t$, the prediction ${\bp}[t]$ is a weighted ensemble of up-sampled $\bp_u$, \ie ${\bp}[t]\!=\!\sum_{u} \alpha_u \cdot \text{up}(\bp_u, T)$,  where $\alpha_u$ is the ensemble weight of decoder $u$ with $\sum \alpha_u\!=\!1$, and $\text{up}\BK{\bp_u, T}$ denotes the upsampled decoder output of length $T$. The sum is action-wise and the final predicted action label is 
$\hat{y}[t] = {\argmax}_{k \in \calA} \, {\bp}[t, k]$. Note that $\bG := \{\bG_u\}$ differs from the evaluation linear classifier $\bG_f$ of~\cref{subsec:linear-evaluation}.  $\bG_f$ is used to evaluate the representation $\mathbf{f}$, whereas $\{\bG_u\}$ is used for semi-supervised learning with only labeled data $\mathcal{D}_L$.

In addition to learning $\bG$, $\calD_L$ can also be leveraged to fine-tune $\mathbf{M}$.
In Eq.~\eqref{eqn:positive-negative}, the positive and negative sets $\calP_{n,i}$ and $\calN_{n,i}$ can be modified for $\calD_L$ to use the ground truth labels by replacing the unsupervised cluster labels $l_n[t^n_i]$ with ground truth action labels $y_n[t^n_i]$. Note that the learning rate used for fine-tuning the parameters of the model $\mathbf{M}$ is significantly lower than the linear projection layers $\bG_u$. The loss used is $\calL = \losscross(\calD_L) + \losscons'(\calD_L)$, where $\losscons'$ is as defined in Eq.~$\eqref{eqn:contrastice-loss}$ but with $l_n[t^n_i]$ replaced by $y_n[t^n_i]$. 


%
%
%
\subsubsection{\textit{Contrast Step: Update $\mathbf{M}$ with $\mathcal{D}_U \cup \mathcal{D}_L$}}\label{subsec:contrast-step}
After fine-tuning, 
$\mathbf{M}$ and $\mathbf{G}$ can be used to
predict frame-level action labels $\hat{y}_n$ for any unlabeled videos, 
\ie pseudo-labels for $\calD_U$. This affords the possibility of 
updating the representation in $\mathbf{M}$. To that end, we again modify $\calP_{n,i}, \calN_{n,i}$ in Eq.~$\eqref{eqn:positive-negative}$ by replacing the cluster labels $l_n[t^n_i]$ with the pseudo-labels $\hat{y}_n[t]$ and ground truth labels $y_n[t^n_i]$ for $\calD_U$ and $\calD_L$ respectively.  $\mathbf{M}$ is then updated by applying the loss $\calL' = \losscons'(\calD_U \cup \calD_L)$,
where $\losscons'$ is as defined in Eq.~$\eqref{eqn:contrastice-loss}$.



\subsubsection{\textit{Iterative-Contrast-Classify} (ICC)}\label{subsec:Iterative-Contrast-Classify}

The pseudo-labels for $\calD_U$ are significantly more representative of the (unseen) action labels than the clusters obtained from the input I3D features used in the unsupervised stage. Thus, we can improve our contrastive representation by using the pseduo-labels (obtained after \textit{classify}) for another \textit{contrast} step. This refined representation can, in turn, help in finding better pseudo-labels through another following \textit{classify} step. By iterating between the contrast and classify steps in ~\cref{subsec:classify-step} and~\cref{subsec:contrast-step} (see ~\cref{fig:ICC-depiction}), we can progressively improve the performance of the semi-supervised segmentation. The segmentation performance is evaluated at the end of the \textit{classify} step after the training of $\mathbf{G}$.  We denote the combined model of $\mathbf{M}$ and $\mathbf{G}$ for each iteration $i$ as $\text{ICC}_{i}$.  In this way, initial unsupervised representation learning can be considered the \emph{``contrast''} step of $\text{ICC}_1$, where cluster labels are used instead of pseudo-labels.  Performance saturates after $4$ iterations of \textit{contrast-classify}; we refer to 
$\text{ICC}_{4}$ as our final semi-supervised result. 


\section{Experiments}\label{sec:results}
\begin{table*}[t]
\begin{center}
\small{
\begin{tabular}{p{3.2cm} | p{0.5cm}p{0.5cm}p{0.45cm}p{0.5cm}p{0.6cm} | p{0.5cm}p{0.5cm}p{0.45cm}p{0.5cm}p{0.6cm} | p{0.5cm}p{0.55cm}p{0.55cm}p{0.5cm}p{0.4cm}}
\hline
& \multicolumn{5}{c|}{Breakfast} & \multicolumn{5}{c|}{50Salads} & \multicolumn{5}{c}{GTEA} \\
\hline
\textbf{Method} & \multicolumn{3}{c}{$F1@\{10,25,50\}$} & Edit & MoF & \multicolumn{3}{c}{$F1@\{10,25,50\}$} & Edit & MoF & \multicolumn{3}{c}{$F1@\{10,25,50\}$} & Edit & MoF\\
\hline
ED-TCN~\cite{TED-lea2017temporal} 
& -- & -- & -- & -- & -- & 
                                68.0 & 63.9 & 52.6 & 52.6 & 64.7 
                                & 72.2 & 69.3 & 56.0 & -- & 64.0\\
TDRN~\cite{TEDresi-lei2018temporal} 
& -- & -- & -- & -- & -- & 
                                \textbf{72.9} & 68.5 & 57.2 & \textbf{66.0} & 68.1 &
                                79.2 & 74.4 & 62.7 & 74.1 & 70.1 \\
 Our Base \textbf{M} 
                & \textbf{59.8} & \textbf{55.6} & \textbf{45.8} & \textbf{60.1} & \textbf{69.3} 
                & 72.1 & \textbf{68.6} & \textbf{57.8} & 63.8 & \textbf{78.5} & 
                \textbf{87.9} & \textbf{86.1} & \textbf{71.6} & \textbf{84.1} & \textbf{77.8} \\ 
\hline
\end{tabular}
}\end{center}
\caption{Our base model \textbf{M} exceeds the performance of existing encoder-decoder TCNs in most metrics for all three datasets.}
\label{tab:base_model_c2f_ensemble}
\end{table*}

\subsection{Experimental Setup} 

\noindent \textbf{Datasets.} 
We evaluate our method on three standard benchmark datasets and additionally on a new challenging segmentation dataset, Assembly101~\cite{sener2022assembly101}. 
{Breakfast Actions}\cite{kuehne2014language} is a third-person view dataset of 1.7k videos with 10 complex activities of making breakfast, with 48 action classes.
{50Salads}~\cite{stein2013combining} features 25 people making 2 mixed salads. There are 50 videos with 19 different action classes.  {GTEA}~\cite{gtea-fathi2011learning} captures 28 egocentric videos with 11 different action classes. Assembly101~\cite{sener2020temporal} contain 4321 videos with 202 different action classes.
\\
\noindent \textbf{Evaluation.} We follow recent works and report Mean-over-frames(MoF), segment-wise edit distance (Edit) and $F1$-scores with IoU thresholds of $0.10$, $0.25$ and $0.50$ ($F1@\{10, 25, 50\}$). For Breakfast, 50Salads and Gtea, we use features pre-computed from a pre-trained Kinetics I3D model ~\cite{carreira2017quo} and follow the dataset-designated cross-validation. For Assembly101~\cite{sener2022assembly101}, we use the dataset's available features from the TSM~\cite{lin2019tsm} model fine-tuned on Epic-Kitchens datasets~\cite{damen2018scaling} and the designated train-val-test splits.
For the semi-supervised framework, we use the specified train-test splits for each dataset and randomly select $5\%$ or $10\%$ of videos from the training split for labeled dataset $\calD_L$. As GTEA and 50Salads are small, we use $3$ and $5$ videos as 5\% and 10\%, respectively, to incorporate all $C$ actions. 
We report mean and standard deviation of five different selections in Appendix D. For unsupervised representation learning, we use all the unlabeled videos in the dataset, which is in line with other unsupervised works~\cite{unsupervised-kukleva2019unsupervised, selfsupervised-chen2020action}.

\noindent \textbf{Implementation Details} The C2F-TCN architecture and choice of hyper-parameters is detailed in the Appendix A. A base window $w_0$ of $\{10, 20, 4, 20\}$ is applied for Breakfast, 50Salads, GTEA and Assembly-101, respectively. The $w_0$'s are chosen to be small enough for no actions to be dropped during down-sampling. 
For all datasets, we use the weights for the coarse-to-fine ensembling of decoder layer i.e $\alpha_u=\frac{1}{n} ,\;\forall u \in [1, \ldots, n]$, where $n$ is number of decoder layers.
\\
For the semi-supervised setup, we sample frames from each video with $K=\{20, 60, 20\}$ partitions,  
{$\epsilon\!\!\approx\!\!\frac{1}{3K}$} for sampling, and temporal proximity $\delta\!=\!\{0.03, 0.5, 0.02\}$ for Breakfast, 50Salads, and GTEA, respectively. 
The contrastive temperature $\tau$ in Eqs.~\eqref{eqn:frame-level-contrast} and \eqref{eqn:video-level-contrast} is set to $0.1$. We also leverage the FA of C2F-TCN in semi-supervised learning. 

\subsection{Evaluation of the Supervised Setup} 

\subsubsection{Detailed Ablation on the Base Encoder-Decoder}

\begin{table*}[t]
\begin{center}
\small{
\begin{tabular}{p{2.9cm} | p{0.5cm}p{0.5cm}p{0.5cm}p{0.5cm}p{0.6cm} | p{0.5cm}p{0.5cm}p{0.5cm}p{0.5cm}p{0.6cm} | p{0.5cm}p{0.5cm}p{0.5cm}p{0.5cm}p{0.6cm}}
\hline
& \multicolumn{5}{c|}{Breakfast} & \multicolumn{5}{c|}{50Salads} & \multicolumn{5}{c}{GTEA} \\
\hline
\textbf{Method} & \multicolumn{3}{c}{$F1@\{10,25,50\}$} & Edit & MoF & \multicolumn{3}{c}{$F1@\{10,25,50\}$} & Edit & MoF & \multicolumn{3}{c}{$F1@\{10,25,50\}$} & Edit & MoF\\
\hline
 without TPP layer $\bottle$ & $70.9$ & $67.6$ & $56.5$ & $67.9$ & $75.5$ &
                $82.7$ & $80.9$ & $71.0$ & $74.5$ & $84.0$ & 
                $90.6$ & $89.3$ & $78.4$ & $87.3$ & $80.4$ \\
 with TPP layer $\bottle$ 
 & \textbf{71.9} & \textbf{68.8} & \textbf{58.5} & \textbf{68.9} & \textbf{76.6} 
 & \textbf{84.3} & \textbf{81.7} & \textbf{72.8} & \textbf{76.3} & \textbf{84.5} 
 & \textbf{92.3} & \textbf{90.1} & \textbf{80.3} & \textbf{88.5} & \textbf{81.2} \\

\hline
\end{tabular}
}\end{center}
\caption{Temporal pyramid pooling (TPP) is the most effective when inputs are of varying resolution. Without a TPP layer from our full stack, \ie{} \modelname{} + FA,  performance is reduced, most notably in F1@50 for 50Salads and GTEA.} \label{tab:bottleneck}
\end{table*}

\begin{table*}[t]
\begin{center}
\small{
\begin{tabular}{p{3.4cm} | p{0.45cm}p{0.45cm}p{0.45cm}p{0.5cm}p{0.6cm} | p{0.5cm}p{0.5cm}p{0.45cm}p{0.5cm}p{0.6cm} | p{0.5cm}p{0.55cm}p{0.55cm}p{0.5cm}p{0.4cm}}
\hline
& \multicolumn{5}{c|}{Breakfast} & \multicolumn{5}{c|}{50Salads} & \multicolumn{5}{c}{GTEA} \\
\hline
\textbf{Method} & \multicolumn{3}{c}{$F1@\{10,25,50\}$} & Edit & MoF & \multicolumn{3}{c}{$F1@\{10,25,50\}$} & Edit & MoF & \multicolumn{3}{c}{$F1@\{10,25,50\}$} & Edit & MoF\\
\hline
*ED-TCN~\cite{TED-lea2017temporal}
                                & 48.1 & 43.6 & 30.9 & 49.2 & 55.3 & 
                                69.2 & 65.3 & 53.4 & 62.7 & 66.5 &
                                75.7 & 72.7 & 60.4 & 76.7 & 66.1\\
\textbf{+}C2F-Ensem 
                                & 49.0 & 44.2 & 31.6 & 49.6 & 56.4 & 
                                70.5 & 66.7 & 53.8 & 63.3 & 68.5 &
                                76.7 & 72.9 & 63.4 & 76.8 & 67.6 \\
\hline
\textbf{C2F-Ensem Gain} & +0.9 & +0.6 & +0.7 & +0.4 & +1.1 &
                                  +1.3 & +1.4 & +0.4 & +0.6 & +2.0 &
                                  +1.0 & +0.2 & 3.0 & +0.1 & +1.5 \\
\hline
Our Base \textbf{M} & 59.8 & 55.6 & 45.8 & 60.1 & 69.3 & 
                72.1 & 68.6	& 57.8 & 63.8 & 78.5 & 
                87.9 & 86.1 & 71.6 & 84.1 & 77.8 \\ 
\textbf{+}C2F-Ensem(\modelname{}) 
& 64.9 & 60.6 & 49.7 & 63.2 & 70.2
                            & 75.6 & 72.7 & 61.2 & 69.1 & 79.6
                            & 89.9 & 88.3 & 75.9 & 86.8 & 79.6  \\
                            \hline
\textbf{C2F-Ensem Gain} & +5.1 & +5.0 & +3.9 & +3.1 & +0.9 
                    & +2.9 & +4.1 & +3.4 & +5.3 & +1.1 
                    & +2.0 & +2.2 & +4.3 & +2.7 & +1.8\\

\hline
\end{tabular}
}\end{center}
\caption{Gain with the C2F-Ensemble in both ED-TCN and our base model \textbf{M}. *ED-TCN is our implementation with $\mathcal{L}_{\text{TR}}$.}
\label{tab:effect_c2f_ensemble}
\end{table*}

We report our base TCN model \textbf{M=(}$\enc{}, \bottle{}, \dec{}$\textbf{)} with the probability outputs $\bp^{(6)}$ from the last decoder layer (\ie without ensembling) in ~\cref{tab:base_model_c2f_ensemble}. Compared to older encoder-decoders like ED-TCN~\cite{TED-lea2017temporal} and  TDRN~\cite{TEDresi-lei2018temporal}, our network is deeper (6 layers vs. 2),
but has approximately the same parameters ($\approx$ 6M) as we use smaller kernels ($5$ vs $25$ for ED-TCN, $50$ for TDRN) and skip connections similar to the Unet~\cite{ronneberger2015u, dac-spp-unet} style architecture. Our designed base encoder-decoder architecture outperformed  ED-TCN~\cite{TED-lea2017temporal} and TDRN~\cite{TEDresi-lei2018temporal} in most metrics on all three datasets (in 
\cref{tab:base_model_c2f_ensemble}).

\begin{figure}
    \centering
    \includegraphics[width=0.7\linewidth]{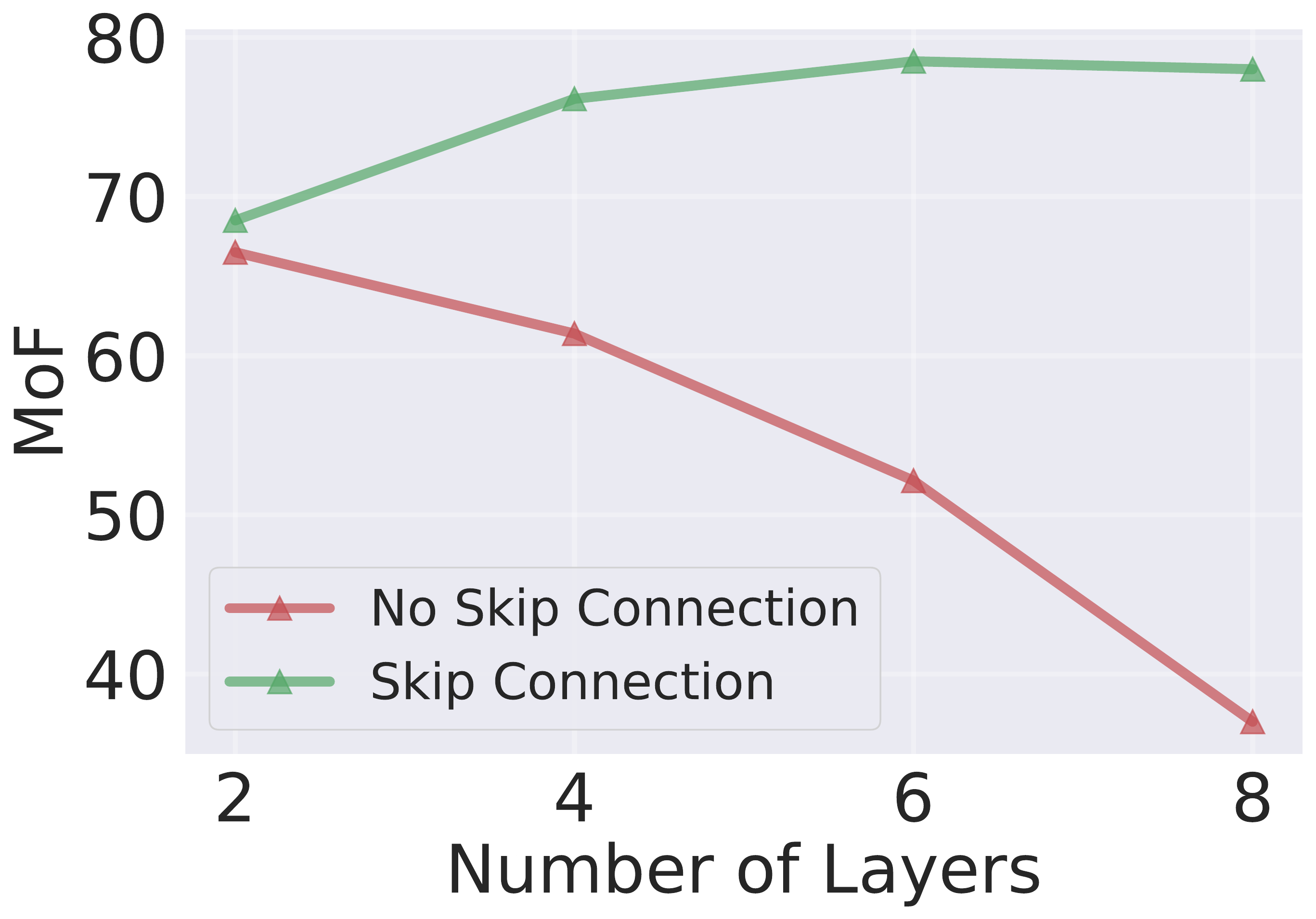}
    \caption{Skip connection ablation on 50Salads: Skip connections help to increase the MoF of the ED-TCN architecture, and its impact is highlighted by the increase in the number of layers of the ED-TCN architecture.}
    \label{fig:skip_connection}
\end{figure}

\noindent \textbf{Relationship between Skip Connection and Number of Layers} As mentioned in ~\cref{sec:base_model}, each decoder layer concatenates its up-sampled outputs with encoder $\enc_{(6 - i)}$'s layers output via \textit{a skip connection}. \cref{fig:skip_connection} visualizes the impact of skip connections with an increase in the number of layers of the ED-TCN (keeping the kernel size fixed at 25). With an increase in the number of layers, the MoF of the ED-TCN drastically decreases. However, with added skip connections (similar to Unet~\cite{ronneberger2015u}) and an increase in the number of layers, the scores increase again. 
The MoF saturates at roughly six layers, which corresponds to our choice. 

\begin{figure}
    \centering
    \includegraphics[width=0.9\linewidth]{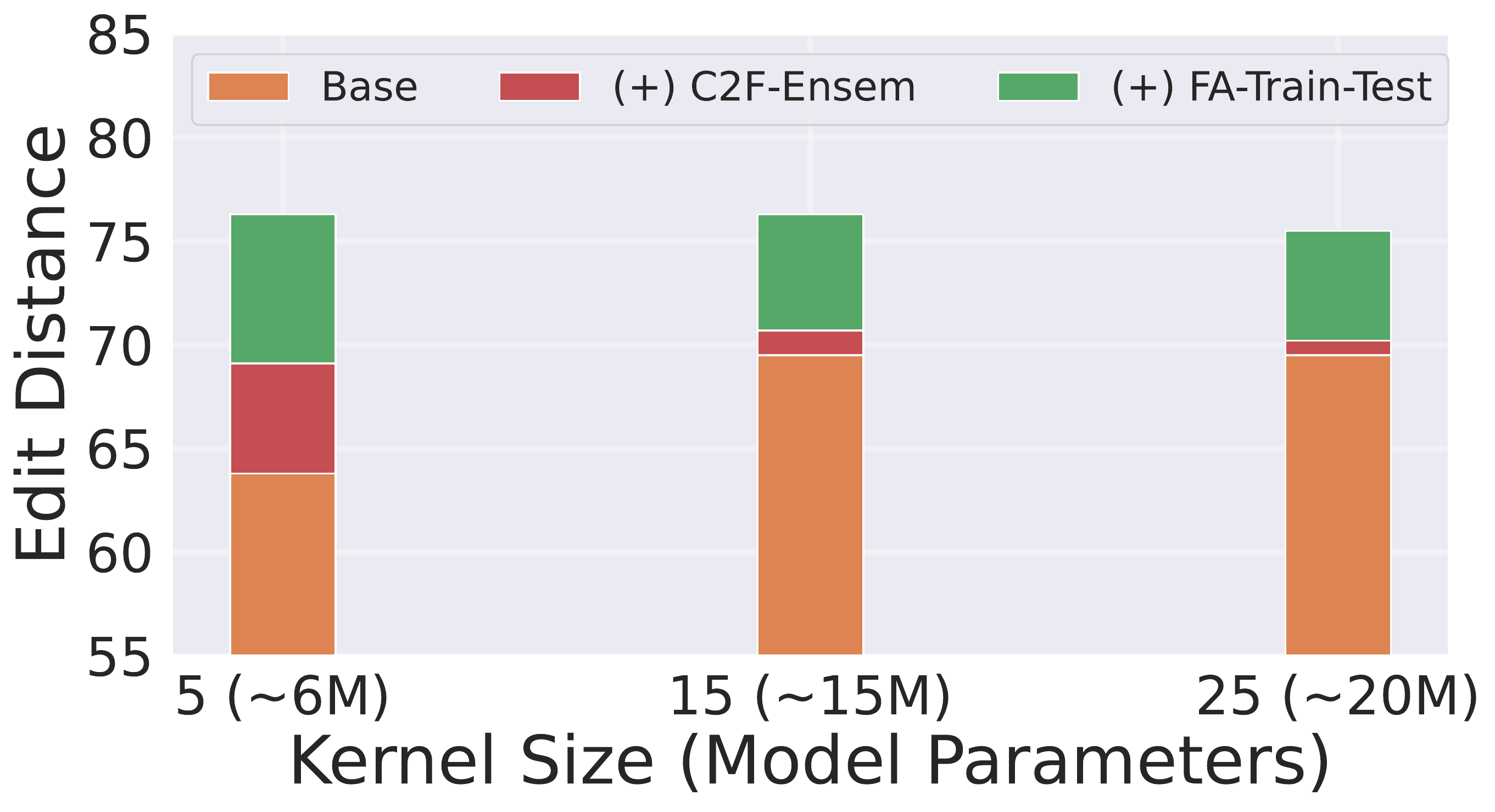}
    \caption{Kernel size ablation on 50Salads: C2F-Ensemble and FA helps in gaining competitive results even with a smaller kernel size. Reducing the convolution kernel size from 25 to 5 within our 6 layered C2F-TCN model leads to a reduction in the parameters from $\approx$ 20M to $\approx$ 6M with similar results.}
    \label{fig:kernel_size}
\end{figure}

\noindent \textbf{Impact of Kernel Size} ~\cref{fig:kernel_size} shows that a convolution kernel of size 5 in the 6-layered C2F-TCN model is sufficient to obtain competitive scores. The base model scores (in orange) increased with larger kernel sizes, saturating at kernel size 25, at the expense of more model parameters. We observe, however, that the C2F ensembling gave the greatest improvement (shown in red) for the TCN with size 5 kernels.  
By incorporating the C2F ensemble, the resulting scores are competitive against the use of larger kernels (and thereby more model parameters). 
As the ensembling requires no additional parameters, yet can improve scores, this again affirms our architecture choice in ensembling. 

\noindent \textbf{Impact of the Bottleneck Layer} \label{sec:impact_of_bottleneck} 
Table \ref{tab:bottleneck} shows the impact of the temporal pooling layer, used to handle multiple temporal resolutions, which is characteristic of video datasets. It is also added through our augmentation strategy. To highlight the importance of including a multi-resolution hidden feature representation at the bottleneck, we show scores after the removal of the bottleneck layer from our final C2F-TCN + FA. The scores decreased in all the datasets and metrics. Interestingly, the biggest impact was seen in the metric F1@50, which is the most strict criteria for over-segmentation.

\begin{table}[]
    \begin{center}
    \small{
    \begin{tabular}{l|c c c c c}
    \hline
    \textbf{Method} & \multicolumn{3}{c}{$F1@\{10,25,50\}$} & Edit & MoF \\
    \hline
        Base ($\enc,\bottle,\dec$) & 59.8 & 55.6 & 45.8 & 60.1 & 69.3 \\
        \textbf{(+)} loss at each decoder & 59.2 & 55.1 & 45.6 & 58.7 & \textbf{70.6} \\
        \textbf{(+)} learned $\alpha_u$ & 61.8 & 57.8 & 47.6 & 61.4 & \textbf{70.5} \\
        \textbf{(+)} fixed $\alpha_u$ (\modelname{})
        & \textbf{64.9} & \textbf{60.6} & \textbf{49.7} & \textbf{63.2} & 70.2 \\
         \hline
    \end{tabular}
    }\end{center}
    \caption{Ablation on ensembling weights on Breakfast.}
    \label{tab:ensemble_vs_loss_every}
\end{table}

\begin{table*}[t]
\begin{center}
\small{
\begin{tabular}{p{3.0cm} | p{0.5cm}p{0.5cm}p{0.45cm}p{0.5cm}p{0.6cm} | p{0.5cm}p{0.5cm}p{0.45cm}p{0.5cm}p{0.6cm} | p{0.5cm}p{0.55cm}p{0.55cm}p{0.5cm}p{0.4cm}}
\hline
& \multicolumn{5}{c|}{Breakfast} & \multicolumn{5}{c|}{50Salads} & \multicolumn{5}{c}{GTEA} \\
\hline
\textbf{Method} & \multicolumn{3}{c}{$F1@\{10,25,50\}$} & Edit & MoF & \multicolumn{3}{c}{$F1@\{10,25,50\}$} & Edit & MoF & \multicolumn{3}{c}{$F1@\{10,25,50\}$} & Edit & MoF\\
\hline     
MS-TCN~\cite{li2020ms} & 64.1 & 58.6 & 45.9 & 65.6 & 67.6 
        & 80.7 & 78.5 & 70.1 & 74.3 & 83.7 
        & 87.8 & 86.2 & 74.4 & 82.6 & 78.9 \\
\textbf{(+)} FA-Train & 70.2 & 66.7 & 56.5 & 67.5 & 70.3 
                   & 81.4 & 79.8 & 72.0 & 76.0 & 84.1
                   & 88.7 & 85.7 & 74.2 & 82.8 & 78.7 \\
\textbf{(+)} FA-Train-Test & 70.8 & 67.6 & 56.8 & 67.7 & 71.3 
                  & 82.8 & 80.4 & 72.2 & 76.1 & 84.2 
                  & 88.8 & 85.7 & 74.1 & 82.8 & 79.0 \\
\hline
\textbf{FA Gain} & +6.7 & +8.9 & +10.9 & +2.1 & +3.7
            & +2.1 & +1.9 & +2.1 & +1.8 & +0.5 
            & +1.0 & -0.5 & -0.3 & +0.2 & +0.1\\
\hline

\modelname{} & 64.9 & 60.6 & 49.7 & 63.2 & 70.2
            & 75.6 & 72.7 & 61.2 & 69.1 & 79.6
            & 89.9 & 88.3 & 75.9 & 86.8 & 79.6 \\
            
\textbf{(+)} FA-Train & 70.8 & 67.5 & 57.3 & 67.5 & 74.3 & 
                        78.9 & 77.1 & 66.9 & 72.5 & 81.9 & 
                        92.2 & 89.9 & 80.2 & 88.0 & 81.2 \\
\textbf{(+)} FA-Train-Test & 71.9 & 68.8 & 58.5 & 68.9 & 76.6 &
                                      84.3 & 81.7 & 72.8 & 76.3 & 84.5 &
                                      92.3 & 90.1 & 80.3 & 88.5 & 81.2 \\
\hline
\textbf{FA Gain} & +7.0 & +8.9 & +8.7 & +5.7 & +6.3 
             & +8.7 & +9.0 & 11.6 & +7.2 & +2.4 
             & +2.4 & +1.8 & +4.4 & +1.7 & +1.6\\
             \hline
 \end{tabular}
 }\end{center}
\caption{Our feature augmentation (FA) strategy significantly improves the performance of both MS-TCN and C2F-TCN.}
\label{tab:feature_augmentation}
\end{table*}

\subsubsection{Impact of the C2F-Ensemble} Instead of adding loss on the last layer, we add loss on the \textit{``coarse-to-fine ensemble''} probabilities (\cref{subsec:multilayer_ensemble}). We show the increment with C2F-Ensemble on all datasets, both with existing ED-TCN (our implementation with added loss $\mathcal{L}_{\text{TR}}$) and with our base model \textbf{M} in \cref{tab:effect_c2f_ensemble}. The increment in ED-TCN is comparatively small because it contains 2 decoder layers, compared to the 6 used here. Ensembling decoder layers on our model \textbf{M} gives a significant improvement in Edit and F1 scores on all three datasets. The ensemble is easy to implement and requires no additional network components. In Appendix B we show that the 
ensembled probabilities of multiple layers in MS-TCN~\cite{li2020ms} do not yield higher results as there is no diversity in the temporal resolution of representation. 

Forcing weights ($\alpha_i$) on lower decoder layers via equal weighting, as done here, performed better than (1) learning $\alpha_i$ and (2) applying individual losses to each decoding layer (see \cref{tab:ensemble_vs_loss_every}). Learned weights 
lead to lower Edit and F1 scores than fixed weights 
because learned weights emphasize latter decoder layers, leading to higher MoF but also higher fragmentation.  

\cref{fig:layer-analysis} shows an analysis of the C2F-Ensemble compared to the individual decoder layers $\dec^{(u)}$. The left plot shows a sample prediction, highlighting that the 
fragmentation errors of the last decoder $\dec^{(6)}$ were mitigated by the earlier layers. 
The plot on the right 
shows that the fourth decoder $\dec^{(4)}$ performed best individually, however, 
the ensemble yielded the highest value for all scores, especially in Edit distance -- the measure most affected by fragmentation.


\subsubsection{Impact of Temporal Feature Augmentation} \cref{tab:feature_augmentation} shows the gains with our augmentation strategy during training \textbf{(FA-Train)} and testing \textbf{(FA-Train-Test)} when applied to MS-TCN~\cite{li2020ms} and our \modelname{}.  FA consistently improved the performance of \modelname{} in all three datasets. Applying FA on top of MS-TCN generally led to better results, most notably on Breakfast, but it was less impressive on GTEA. GTEA has marginal decreases in F1; we speculate that the short segments, a characteristic of GTEA, became completely lost in the coarser windows. On the larger Breakfast and 50Salads datasets, MS-TCN with FA outperformed in most metrics compared to BCN/GatedR \ie MS-TCN with additional refinement modules (shown in~\cref{tab:SOTA}).

\begin{table*}[t]
\begin{center}
\small{
\begin{tabular}{p{3.0cm} | p{0.5cm}p{0.5cm}p{0.45cm}p{0.5cm}p{0.6cm} | p{0.5cm}p{0.5cm}p{0.45cm}p{0.5cm}p{0.6cm} | p{0.5cm}p{0.55cm}p{0.55cm}p{0.5cm}p{0.4cm}}
\hline
& \multicolumn{5}{c|}{Breakfast} & \multicolumn{5}{c|}{50Salads} & \multicolumn{5}{c}{GTEA} \\
\hline
\textbf{Method} & \multicolumn{3}{c}{$F1@\{10,25,50\}$} & Edit & MoF & \multicolumn{3}{c}{$F1@\{10,25,50\}$} & Edit & MoF & \multicolumn{3}{c}{$F1@\{10,25,50\}$} & Edit & MoF\\
\hline
MS-TCN~\cite{li2020ms} & 64.1 & 58.6 & 45.9 & 65.6 & 67.6 & 
                                            80.7 & 78.5 & 70.1 & 74.3 & 83.7 & 
                                            87.8 & 86.2 & 74.4 & 82.6 & 78.9 \\

GatedR~\cite{wang2020gated} & 71.1 & 65.7 & 53.6 & 70.6 & 67.7 & 78.0 & 76.2 & 67.0 & 71.4 & 80.7 & 89.1 & 87.5 & 72.8 & 83.5 & 76.7 \\
BCN~\cite{wang2020boundary} & 68.7 & 65.5 & 55.0 & 66.2 & 70.4 & 82.3 & 81.3 & \textbf{74.0} & 74.3 & 84.4 & 88.5 & 87.1 & 77.3 &	84.4 & 79.8 \\
DTGRM~\cite{wang2020temporal} & 57.5 & 54.0 & 43.3 & 58.7 & 65.0 & 75.4 & 72.8 & 63.9 & 67.5 & 82.6  & -- & -- & -- & -- & -- \\
G2L~\cite{gao2021global2local} & \textbf{74.9} & \textbf{69.0} & 55.2 & \textbf{73.3} & 70.7
              & 80.3 & 78.0 & 69.8 & 73.4 & 82.2 
              & 89.9 & 87.3 & 75.8 & 84.6 & 78.5\\

\hline
\textbf{\modelname{}} & 64.9 & 60.6 & 49.7 & 63.2 & 70.2 &  
                    75.6 & 72.7 & 61.2 & 69.1 & 79.6 &
                    89.9 & 88.3 & 75.9 & 86.8 & 79.6 \\
\textbf{MS-TCN} (+) \textbf{FA} 
& 70.8 & 67.6 & 56.8 & 67.7 & 71.3 &                           82.8 & 80.4 & 72.2 & 76.1 & 84.2 & 
  88.8 & 85.7 & 74.1 & 82.8 & 79.0\\
\textbf{\modelname{}} (+) \textbf{FA} 
& 71.9 & \textbf{69.0} & \textbf{58.5} & 68.9 & \textbf{76.6} &
\textbf{84.3} & \textbf{81.7} & 72.8 & \textbf{76.3} & \textbf{84.5} &
\textbf{92.3} & \textbf{90.1} & \textbf{80.3} & \textbf{88.5} & \textbf{81.2} \\
\hline

\end{tabular}
}\end{center}
\caption{State-of-the-art comparisons verifying that \modelname{} with our feature augmentation (FA) exceeds other methods in most metrics in all datasets. We obtain noteworthy gains of +3.3 F1@50, +5.9 MoF on the largest Breakfast dataset and significantly exceed other methods on 50Salads and GTEA.}\label{tab:SOTA}
\end{table*}

\begin{table}[]
    \begin{center}
    \small{
    \begin{tabular}{l|ccccc}
    \hline
    Method & \multicolumn{3}{c}{$F1@\{10,25,50\}$} & Edit & MF \\
    \hline
    MS-TCN~\cite{li2020ms} & 17.1 & 14.1 & 8.7 & 21.0 & 21.2 \\
    C2F-TCN & \textbf{20.2} & \textbf{16.6} & \textbf{10.8} & \textbf{22.3} & \textbf{22.5} \\
    \hline
    \end{tabular}
    }
    \end{center}
    \caption{C2F-TCN outperforms MS-TCN in the task of coarse segmentation on the challenging Assembly101 dataset.}
    \label{tab:Assembly101-coarse}
\end{table}

\begin{figure*}
\begin{center}
\includegraphics[width=0.9\linewidth]{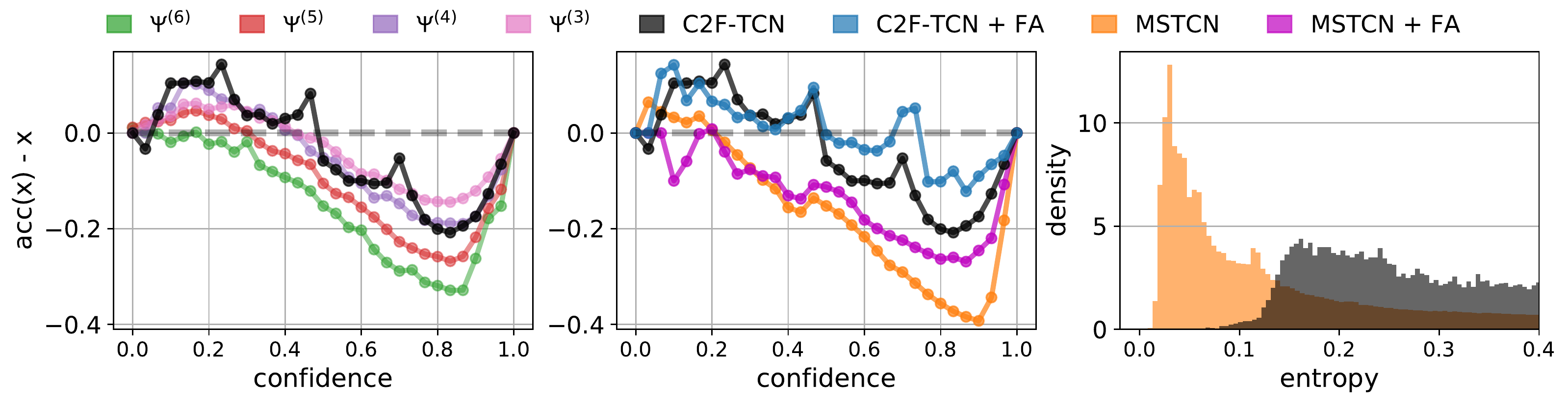}
\end{center}
\caption{Uncertainty quantification: Calibration curves
analyzing our ensemble (left) and showing the impact of feature augmentation (middle).  The ideal curve sits at 0; values above and below indicate under-/over-confidence, respectively. Our ensemble is better calibrated and less over-confident than the final decoder layers as well as MS-TCN. Adding our feature augmentation strategy `\textit{(+FA)}' further improves the calibration of our own model and MS-TCN. The rightmost plot is the density of the entropy of probability for incorrect predictions. Our C2F ensemble is more uncertain about wrong predictions than MS-TCN.}
\label{fig:uncertainty}
\end{figure*}

\subsubsection{Comparison with \textit{State-of-the-Art}}\label{subsec:sota_comparison}
\cref{tab:SOTA} compares our results with the most recent segmentation works, MS-TCN~\cite{li2020ms}, GatedR~\cite{wang2020gated}, BCN~\cite{wang2020boundary}, GTRM~\cite{wang2020temporal}, and Global2Local(G2L)~\cite{gao2021global2local}. All listed works use $I3D$ features.  We omit works not directly comparable, such as~\cite{ishikawa2021alleviating}, which  uses features extracted from an already-trained MS-TCN, and~\cite{selfsupervised-chen2020action}, which accesses (unlabeled) test videos during training.  
BCN, GatedR and G2L are built on top of MS-TCN, a feedforward TCN with fixed temporal resolution. DGTRM is built on graph convolutional networks. ED-TCN and our C2F-TCN are encoder-decoder architectures that down- and then up-sample in time. 

Our model outperformed the \textit{state-of-the-art} scores by +5.9\% and +3.3\% on the MoF and F1@50, respectively, on Breakfast, the largest of the three datasets. G2L, formed using neural architecture search to improve MS-TCN's receptive field, had a slightly higher performance than our method in Breakfast Edit and F1@10 scores; however, we exceeded it in all other metrics and datasets. For 50Salads, we outperformed by +2.0 \% on Edit and F1@10.  For F1@50, we were slightly lower than BCN, but BCN is worse for all other metrics and datasets. On GTEA, we outperformed the state-of-the-art scores on all metrics by large margins.
From these strong scores, we conclude that \modelname{} generalizes well to different datasets.

In ~\cref{tab:Assembly101-coarse}, we apply MS-TCN~\cite{li2020ms} and C2F-TCN for a coarse-action segmentation task on the Assembly101 dataset and show that C2F-TCN outperforms MS-TCN.

\subsection{Evaluation of Complex Activity Recognition} 

\begin{table}
\begin{center}
\small{
\begin{tabular}{p{4.5cm}|c|c} 
\hline
 \multirow{2}{*}{Method} 
& I3D & Fine-tuned \\
 & features &  features\\
\hline
Timeception~\cite{highlevel-hussein2019timeception} 
& 71.3 & 86.9 \\
PIC~\cite{highlevel-hussein2020pic}  
& - & 89.9 \\
Actor-Focus~\cite{highlevel-ballan2021long} & 72.0 
& 89.9 \\
\hline
Ours Encoder($\enc$) Model 
& 92.4 & -\\
\textbf{(+)} FA-Train 
& $\textbf{94.6}$ & - \\
\hline
margin wrt SOTA & +22.6 & +4.7 \\ 
\hline
\end{tabular}
}\end{center}
\caption{Complex Activity Recognition on the Breakfast dataset: The encoder of C2F-TCN with feature augmentation outperforms previous state-of-the-art even without fine-tuned features.}
\label{tab:high-level-action-results}
\end{table}

\cref{tab:high-level-action-results} compares various dedicated frameworks, namely Timeception~\cite{highlevel-hussein2019timeception}, PIC~\cite{highlevel-hussein2020pic}, and Actor-Focus~\cite{highlevel-ballan2021long},
on Breakfast. To ensure a fair comparison, the same
$1357\!\!:\!\!335$ 
train-test split from Timeception is used. We employ Kinetics pre-trained I3D features that are not fine-tuned on Breakfast, while previous works report additional results with fine-tuning.  
Our base encoder $\enc$ model was +20.9\% above the other methods that do not use fine-tuned features and +2.5\% above those using fine-tuned features. Adding our FA and without using fine-tuned features, we exceeded the \textit{state-of-the-art} results derived using fine-tuned features by +4.7\%.

\subsection{Uncertainty Quantification}\label{subsec:uncertainty_quantification}
Using calibration notations from ~\cref{subsec:calibration_notation}, we partition the confidence values into $N$ equal length bins $\calP_n := (\frac{n}{N}, \frac{n+1}{N}]$ and plot the difference between the associated accuracy and the confidence, \ie $\acc(\calP_n) - \conf(\calP_n)$.  Perfectly calibrated outputs would sit at $0$; over or under $0$ indicates under-/over-confidence. 
The left plot of ~\cref{fig:uncertainty} shows that each decoding layer became progressively more over-confident, especially at higher confidence values. Ensembling the results generally results in calibration levels similar to earlier decoders $\dec^{(3)}$ and $\dec^{(4)}$, while achieving much higher accuracy (see ~\cref{fig:layer-analysis}). The middle plot shows that our FA improved the calibration of both C2F-TCN and MS-TCN. Thus, C2F-Ensemble and FA are two efficient ways to improve the calibration of segmentation models. Calibration is important for real-life application, \ie{} the model must not have highly confident wrong predictions. To show this effect,  
the last plot calculates the \textit{Shannon entropy} of probability predictions for the \textbf{incorrectly classified frames} of all test videos. Higher entropy indicates more uncertainty in prediction. We plot the entropy density for MS-TCN and C2F-TCN. C2F-TCN was more uncertain in the wrong predictions density plot and thus more calibrated.

\begin{table*}
\begin{center}
\small{
\begin{tabular}{l|ccccc|ccccc|ccccc}
    \hline
    & \multicolumn{5}{c|}{Breakfast} & \multicolumn{5}{c|}{50Salads} &  \multicolumn{5}{c}{GTEA}\\\hline
     & \multicolumn{3}{c}{$F1@\{10,25,50\}$} & Edit & MF & \multicolumn{3}{c}{$F1@\{10,25,50\}$} & Edit & MF &
     \multicolumn{3}{c}{$F1@\{10,25,50\}$} & Edit & MF \\
    \hline
    Input I3D Baseline & 4.9 & 2.5 & 0.9 & 5.3 & 30.2 & 12.2 & 7.9 & 4.0 & 8.4 & 55.0 & 48.5 & 42.2 & 26.4 & 40.2 & 61.9 \\
    Our Representations &\textbf{57.0} & \textbf{51.7} & \textbf{39.1} & \textbf{51.3} & \textbf{70.5}
     & \textbf{40.8} & \textbf{36.2} & \textbf{28.1} & \textbf{32.4} & \textbf{62.5} 
     & \textbf{70.8} & \textbf{65.0} & \textbf{48.0} & \textbf{65.7} & \textbf{69.1} \\
    \hline
    Improvement & 52.1 & 49.2 & 38.2 & 46.0 & 40.3 
                & 28.6 & 28.3 & 24.1 & 24.0 & 7.5 
                & 22.3 & 22.8 & 21.6 & 25.5 & 7.2 \\
    \hline
    
\end{tabular}
} 
\end{center}
\caption{Our unsupervised learning represents a large improvement in segmentation compared to input features.}\label{tab:improvement_unsupervised_representation}
\end{table*}

\subsection{Evaluation of C2F-TCN on Representation Learning}\label{subsec:eval-learned-repr}

\begin{table*}[t]
\begin{center}
\small{
\begin{tabular}{c|ccccc|ccccc|ccccc}
\hline
& \multicolumn{5}{c|}{Breakfast} & \multicolumn{5}{c|}{50Salads} & \multicolumn{5}{c}{GTEA} \\
\hline
& \multicolumn{3}{c}{$F1@\{10,25,50\}$} & Edit & MoF & \multicolumn{3}{c}{$F1@\{10,25,50\}$} & Edit & MoF & \multicolumn{3}{c}{$F1@\{10,25,50\}$} & Edit & MoF \\
\hline
    Cluster & 11.7 & 8.0 & 3.9 & 12.2 & 36.1 
                      & 18.5 & 13.7 & 8.5 & 13.6 & 50.8
                      & 57.3 & 48.6 & 31.6 & 52.4 & 60.5\\
    \textbf{(+)} Proximity & 24.4 & 19.2 & 11.5 & 21.3 & 50.0
                                & 18.6 & 13.5 & 8.0 & 13.5 & 51.6 
                                & 62.9 & 56.6 & 38.0 & 52.6 & 62.2\\
    \textbf{(+)} Video-Level & 42.9 & 37.6 & 26.6 & 36.4 & 66.1  
                              & -- & -- & -- & -- & -- 
                              & -- & -- & -- & -- & -- \\
    \hline

\end{tabular}
} 
\end{center}
\caption{Contribution of clustering and time-proximity conditions and video-level constraints for contrastive learning (with $\bz_6$).}\label{tab:ablation_unsupervised_representation}
\end{table*}

\begin{table*}[t]
\begin{center}
\small{
\begin{tabular}{c|ccccc|ccccc|ccccc}
\hline
& \multicolumn{5}{c|}{Breakfast} & \multicolumn{5}{c|}{50Salads} & \multicolumn{5}{c}{GTEA} \\
\hline
 & \multicolumn{3}{c}{$F1@\{10,25,50\}$} & Edit & MoF & \multicolumn{3}{c}{$F1@\{10,25,50\}$} & Edit & MoF & \multicolumn{3}{c}{$F1@\{10,25,50\}$} & Edit & MoF \\
\hline
    Last-Layer($\bz_6$) &  42.9 & 37.6 & 26.6 & 36.4 & 66.1 
                              & 18.6 & 13.5 & 8.0 & 13.5 & 51.6
                              & 62.9 & 56.6 & 38.0 & 52.6 & 62.2\\
    Multi-Resolution($\feat$) & \textbf{57.0} & \textbf{51.7} & \textbf{39.1} & \textbf{51.3} & \textbf{70.5}
     & \textbf{40.8} & \textbf{36.2} & \textbf{28.1} & \textbf{32.4} & \textbf{62.5} 
     & \textbf{70.8} & \textbf{65.0} & \textbf{48.0} & \textbf{65.7} & \textbf{69.1} \\
    
    \hline
    Improvement & 14.1 & 14.1 & 12.5 & 14.9 & 4.4 
                & 22.2 & 22.7 & 20.1 & 18.9 & 10.9 
                & 7.9 & 8.4 & 10.0 & 13.1 & 6.9 \\
    \hline
    \end{tabular}
    }\end{center}
    \caption{Using Multi-Resolution($\feat$) representation instead of a final decoder $\bz_6$ significantly improves the learned representation scores.}
    \label{tab:multi-resolution-improvement}
\end{table*}

\subsubsection{Linear Classification Accuracy} ~\cref{tab:improvement_unsupervised_representation} shows our unsupervised representation learning (see ~\cref{subsec:linear-evaluation}) results.  We evaluate the input I3D features with a linear evaluation protocol to serve as a baseline. Our representation brings significant gains over the input I3D, verifying the ability of the base TCN to perform the task of segmentation with our designed unsupervised learning.



\subsubsection{Frame- and Video-Level Contrastive Learning} ~\cref{tab:ablation_unsupervised_representation} breaks down the contributions from ~\cref{sec:frame_level_contratsive} and ~\cref{subsec:video-level-contrast} when forming the positive and negative sets of contrastive learning from Eq.~$\eqref{eqn:positive-negative}$. The `\textit{Cluster}' row applies the cluster labels condition \ie $l_n[t^n_i] = l_m[t^m_j]$ and `\textit{ (+) Proximity}' adds the condition $|t^n_i - t^m_j| < \delta$.
Adding time proximity was more effective for Breakfast and GTEA, likely because their videos follow a more rigid sequencing than 50Salads. Adding the \textit{Video-Level} contrastive loss from ~\cref{subsec:video-level-contrast} in Breakfast gave a further boost. 

\subsubsection{Multi-resolution representation} ~\cref{tab:multi-resolution-improvement} verifies that our multi-resolution representation $\feat$ (see ~\cref{subsec:multi-resolution-similarity}) outperformed the use of only the final decoder layer feature $\bz_6$ by very large margins.  Gains are especially notable for the F1 score and Edit distance, verifying that $\mathbf{f}$ has less over-segmentation.

\subsection{Evaluation of Semi-Supervised Learning}

\subsubsection{ICC Components}~\cref{tab:icc_improvement} shows the progressive improvements as we increase the number of iterations of our proposed ICC algorithm. The gain in performance was especially noticeable for the Edit and F1 scores. 
The reported segmentation results are from after the \textit{classify} step.  The improvements gained by updating the feature representation after the \textit{contrast} step but before the \textit{classify} step of the next iteration are shown in Appendix C.

\begin{table*}[t]
\begin{center}
\small{
\begin{tabular}{c|c|ccccc|ccccc|ccccc}
\hline
& & \multicolumn{5}{c|}{Breakfast} & \multicolumn{5}{c|}{50Salads} & \multicolumn{5}{c}{GTEA} \\
\hline
&\textbf{Method} & \multicolumn{3}{c}{$F1@\{10,25,50\}$} & Edit & MoF & \multicolumn{3}{c}{$F1@\{10,25,50\}$} & Edit & MoF & \multicolumn{3}{c}{$F1@\{10,25,50\}$} & Edit & MoF \\
\hline
\multirow{4}{*}{\textbf{$\approx$5}} & $\text{ICC}_1$ 
                    & 54.5 & 48.7 & 33.3 & 54.6 & 64.2 
                    & 41.3 & 37.2 & 27.8 & 35.4 & 57.3
                    & 70.3 & 66.5 & 49.5 & 64.7 & 66.0 \\
& $\text{ICC}_2$  
                    & 56.9 & 51.9 & 34.8 & 56.5 & 65.4 
                    & 45.7 & 40.9 & 30.7 & 40.9 & 59.5
                    & 77.0 & 70.6 & 54.1 & 67.8 & 68.0\\
& $\text{ICC}_3$ 
                    & 59.9 & 53.3 & 35.5 & 56.3 & 64.2 
                    & 50.1 & 46.7 & 35.3 & 43.7 & 60.9
                    & 77.6 & 71.2 & 54.2 & 71.3 & 68.0\\
& $\text{ICC}_4$  
                    & 60.2 & 53.5 & 35.6 & 56.6 & 65.3
                    & 52.9 & 49.0 & 36.6 & 45.6 & 61.3 
                    & 77.9 & 71.6 & 54.6 & 71.4 & 68.2 \\
                    
\cline{2-17}
& Gain 
                    & 5.7 & 4.8 & 2.3 & 2.0 & 1.1
                    & 11.6 & 11.8 & 8.8 & 10.2 & 4.0
                    & 7.6 & 5.1 & 5.1 & 6.7 & 2.2 \\

\hline
\end{tabular}
} 
\end{center}
\caption{{Progressive semi-supervised improvement with more iterations of ICC on the 3 benchmark datasets with 5\% labeled training data.}}\label{tab:icc_improvement}
\end{table*}

\begin{table*}[t]
\begin{center}
\small{
\begin{tabular}{c|c|ccccc|ccccc|ccccc}
\hline
& & \multicolumn{5}{c|}{Breakfast} & \multicolumn{5}{c|}{50Salads} & \multicolumn{5}{c}{GTEA} \\
\hline
\%$D_L$ &\textbf{Method} & \multicolumn{3}{c}{$F1@\{10,25,50\}$} & Edit & MoF & \multicolumn{3}{c}{$F1@\{10,25,50\}$} & Edit & MoF & \multicolumn{3}{c}{$F1@\{10,25,50\}$} & Edit & MoF \\
\hline
\multirow{3}{*}{\textbf{$\approx$5}} & Supervised 
                    & 15.7 & 11.8 & 5.9 & 19.8 & 26.0 
                    & 30.5 & 25.4 & 17.3 & 26.3 & 43.1
                    & 64.9 & 57.5 & 40.8 & 59.2 & 59.7\\

& Semi-Super 
                    & \textbf{60.2} & \textbf{53.5} & \textbf{35.6} & \textbf{56.6} & \textbf{65.3}
                    & \textbf{52.9} & \textbf{49.0} & \textbf{36.6} & \textbf{45.6} & \textbf{61.3} 
                    & \textbf{77.9} & \textbf{71.6} & \textbf{54.6} & \textbf{71.4} & \textbf{68.2} \\
\cline{2-17}
& Gain 
                    & 44.5 & 41.7 & 29.7 & 36.8 & 39.3 
                    & 22.4 & 23.6 & 19.3 & 19.3 & 18.2
                    & 13.0 & 14.1 & 13.8 & 12.2 & 8.5
                    \\
\hline
\multirow{3}{*}{\textbf{$\approx$10}} & Supervised 
        & 35.1 & 30.6 & 19.5 & 36.3 & 40.3 
        & 45.1 & 38.3 & 26.4 & 38.2 & 54.8 
        & 66.2 & 61.7 & 45.2 & 62.5 & 60.6 \\ 

& Semi-Super 
        & \textbf{64.6} & \textbf{59.0} & \textbf{42.2} & \textbf{61.9} & \textbf{68.8} 
        & \textbf{67.3} & \textbf{64.9} & \textbf{49.2} & \textbf{56.9} & \textbf{68.6} 
        & \textbf{83.7} & \textbf{81.9} & \textbf{66.6} & \textbf{76.4} & \textbf{73.3} \\
\cline{2-17}
& Gain
        & 29.5 & 28.4 & 22.7 & 25.6 & 28.5 
        & 22.2 & 26.6 & 22.8 & 18.7 & 13.8 
        & 17.5 & 20.2 & 21.4 & 13.9 & 12.7\\

\hline
        

\textbf{100} & Supervised* & 70.8 & 67.5 & 57.3 & 67.5 & 74.3 & 
                        78.9 & 77.1 & 66.9 & 72.5 & 81.9 & 
                        92.2 & 89.9 & 80.2 & 88.0 & 81.2 \\
\hline
\end{tabular}
} 
\end{center}
\caption{Our final all-metrics evaluation of the proposed ICC algorithm on 3 benchmark action segmentation datasets. Semi-Super (our $\text{ICC}_4$) significantly outperforms a supervised counterpart using the same labeled data amount.  See also ~\cref{fig:teaser_semi}. *our C2F-TCN without test-augmentation. }\label{tab:final_semi_supervised_results}
\end{table*}

\subsubsection{Semi-Supervised vs Supervised}
\begin{figure}[t]
\centering
\includegraphics[width=0.9\linewidth]{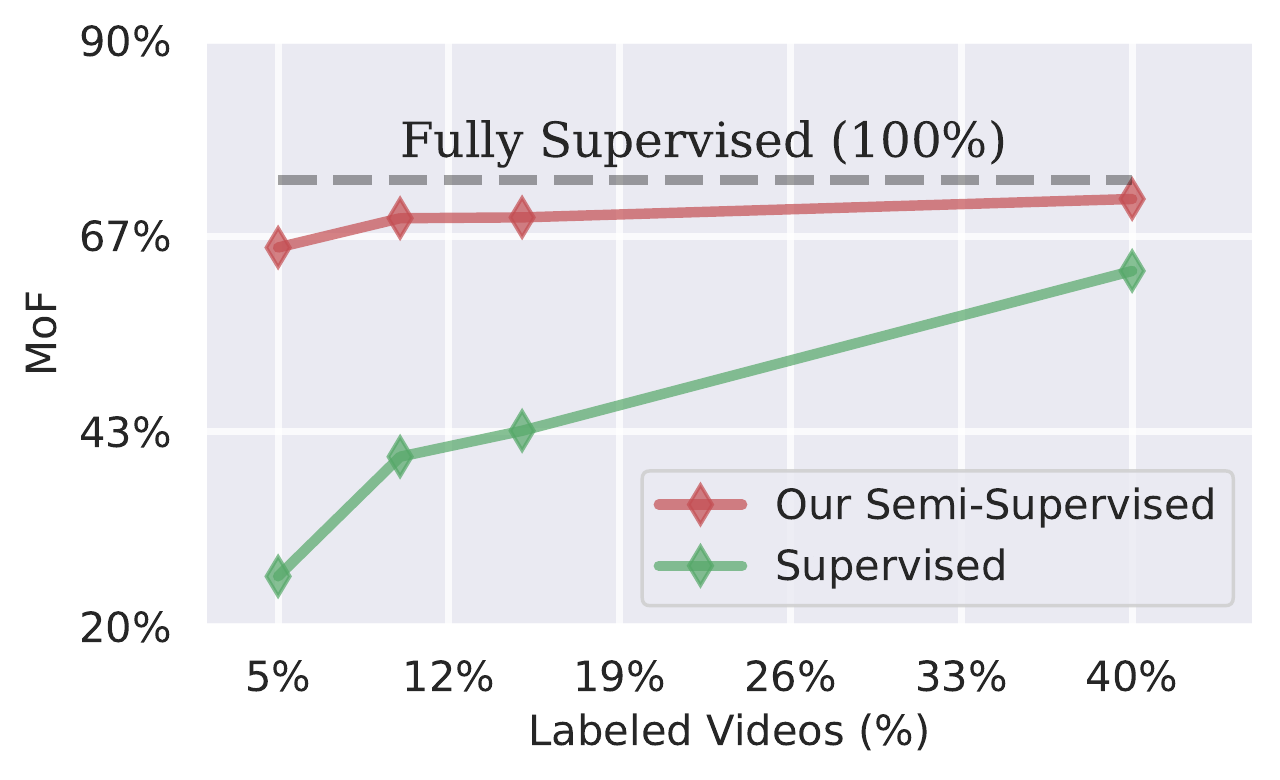}
\caption{Our semi-supervised approach on the Breakfast dataset shows impressive performance with just 5\% labeled videos; at 40\%, we almost match the MoF of a 100\% fully supervised setup.
}
\label{fig:teaser_semi}
\end{figure}

~\cref{tab:final_semi_supervised_results} shows 
our final \emph{`Semi-Super'} results, \ie $\text{ICC}_4$, for various percentages of labeled data.  We compare the \textit{`Supervised'} case of training the base model C2F-TCN with the same labeled dataset $\calD_L$;
ICC significantly outperforms the supervised counterparts' baselines for all metrics (see also ~\cref{fig:teaser_semi}) and for all amounts of training data.
The 100\% supervised C2F-TCN results are reported without test-time augmentations. 
In fact, with just 5\% of labeled videos, there is only 9\% less in MoF in the Breakfast actions compared to fully supervised (100\%). Using less than 5\% (3 videos for 50Salads and GTEA) for training videos does not ensure coverage of all the actions.

\subsubsection{Comparison with 
\textit{State-of-the-Art}} As the first to perform semi-supervised temporal action segmentation, our work is not directly comparable with other works. Table~\ref{tab:different_supervsion_breakfast} shows that our MoF is competitive with other forms of supervision on all three datasets. 
TSS and SSTDA uses weak labels for \textit{all} training videos, while our work requires full labels for only few training videos.  

\begin{table}
\begin{center}
\small{
    \begin{tabular}{c|c|ccc}
    \hline
     & Method & Breakfast & 50Salads & GTEA \\\hline
      & MS-TCN~\cite{li2020ms} & 67.6 & 83.7 & 78.9 \\
     & SSTDA~\cite{selfsupervised-chen2020action}  & 70.2 & 83.2 & 79.8 \\
     \multirow{-3}{*}{\textbf{Full}} & *C2F-TCN  & 74.3 & 81.9 & 81.2 \\
    \hline
     & SSTDA(65\%)~\cite{selfsupervised-chen2020action}  & 65.8 & 80.7 & 75.7\\
      \multirow{-2}{*}{\textbf{Weakly}} & TSS~\cite{timestamp-weakly-li2021temporal} & 64.1 & 75.6 & 66.4 \\
    \hline
     & {Ours ICC (40\%)} & 71.1 & 78.0 & 78.4 \\
     & {Ours ICC (10\%)} & 68.8 & 68.6 & 73.3 \\
      \multirow{-3}{*}{\textbf{Semi}} & {Ours ICC (5\%)}  & 65.3 & 61.3 & 68.2 \\\hline
    \end{tabular}
}\end{center}
\caption{Our semi-supervised results are competitive against different supervision levels. With 40\% labeled training videos, we are close in terms of MoF with fully supervised C2F-TCN counterpart.* indicates with test augmentation.}
\label{tab:different_supervsion_breakfast}
\end{table}

\begin{table}[]
    \begin{center}
    \small{
    \begin{tabular}{l|l|ccccc}
    \hline
     \%$\calD_L$ & \textbf{Method} & \multicolumn{3}{c}{$F1@\{10,25,50\}$} & Edit & MoF\\\hline
        100\% & *Full-Supervised & 70.5 & 66.7 & 53.8 & 63.3 & 68.5 \\
        \hline
         5\% & Supervised &  32.4 & 26.5 & 14.8 & 25.5 & 39.1 \\
         5\% & our ICC &  39.3 & 34.4 & 21.6 & 32.7 & 46.4 \\
         \cline{2-7}
         5\% & Gain & 6.9 & 7.9 & 6.8 & 7.2 & 7.3 \\
         \hline
    \end{tabular}
    }\end{center}
    \caption{Our semi-supervised (final $\text{ICC}_4$) results with ED-TCN~\cite{TED-lea2017temporal} on 50salads with 5\% labeled data significantly improves over its supervised counterpart.* indicates our implementation from ~\cref{tab:effect_c2f_ensemble}} \label{tab:ED-TCN_semi_super}
\end{table}

\subsubsection{Other Baseline TCN Models}




We try our entire semi-supervised ICC algorithm~\cref{fig:ICC-depiction}
using two other base TCN models: ED-TCN (an encoder-decoder architecture) and MS-TCN (a wavenet-like refinement architecture).

\textbf{ED-TCN:} We show that our proposed ICC works with ED-TCN~\cite{TED-lea2017temporal} in~\cref{tab:ED-TCN_semi_super}, whereby ICC showed improved performance over its supervised counterpart. Due to the smaller capacity of ED-TCN compared to C2F-TCN (as indicated by the fully supervised performance of ED-TCN 68.5\% vs C2F-TCN 79.6\% MoF), the ICC algorithm improvement with ED-TCN was lower 
compared to
C2F-TCN. Model capacity influence on representation learning is in line with findings shown in SimCLR~\cite{simCLR} constrastive framework.

\textbf{MS-TCN:} Our proposed unsupervised representation learning did not work well with MS-TCN~\cite{li2020ms}. This was possibly due to the fact that MS-TCN is not designed for representation learning as 3 out of the 4 model blocks consist of refinement stages, whereby each stage takes \textit{class probability} vectors as input from the previous stages. Therefore, representation learning and classifier cannot be decoupled, making alternative classifier-representation learning algorithms difficult. Further, MS-TCN does not have multiple temporal resolution representation such as the encoder-decoder architecture, which plays a significant role in our contrastive learning, as discussed earlier.

\section{Conclusion}
In this work, we designed a temporal encoder-decoder model C2F-TCN 
combining both the coarse and fine decoder outputs of different temporal resolutions. 
In the supervised setting, we performed a coarse-to-fine ensemble of the predictions from the decoding layers, which achieved state-of-the-art performance in supervised temporal action segmentation. 
Ensembled representation produced calibrated predictions with better uncertainty measures, which is crucial for real-world deployment. 
Additionally, the encoder of our C2F-TCN architecture 
achieved state-of-the-art performance in activity recognition, indicating the generalization capability of the model.
In the unsupervised feature learning framework, we formed multi-resolution representation implicitly with outputs from multiple decoder layers of C2F-TCN, bringing temporal continuity, and consequently large improvements in contrastive representation learning. 
%
We showed and utilized the fact that the pre-trained input features that capture the semantics and motion of short-trimmed video segments can be used to learn higher-level representations to interpret long video sequences. Our final iterative semi-supervised learning algorithm ICC can significantly reduce the annotation efforts, with 40\% labeled videos approximately achieving fully supervised (100\%) performance.


As a future direction, we would also like to extend our framework to a combined semi and weak supervision setup. Moreover, we would like to explore using the framework for video domain adaptation, as it is strongly linked with unsupervised contrastive feature learning. 

Although our proposed methods are extremely strong, like previous temporal action segmentation works, our setup also utilizes pre-computed features to avoid the computation expenses of end-to-end training. This serves as a motivation for us to design less computationally expensive end-to-end learning frameworks. Furthermore, similar to previous state-of-the-art TCNs, C2F-TCN also requires the entire video to be available for inference. Hence, the scope is limited to offline temporal action segmentation.

\appendices
\renewcommand{\thetable}{T\arabic{table}}
\renewcommand{\thefigure}{F\arabic{figure}}
\renewcommand{\thesection}{S\arabic{section}}



\section{Details of the C2F-TCN Architecture}\label{sec:model_details}
The following presents the detailed model architecture explained in section 3.1. To define the model, we first define a block called the \textit{double\_conv} block, where \textit{double\_conv(in\_c, out\_c)} = \textit{\textbf{Conv1D(in\_c, out\_c, kernel=5, pad=1)}} $\xrightarrow[]{}$ \textit{BatchNorm1D(out\_c)} $\xrightarrow[]{}$ \textit{ReLU()} $\xrightarrow[]{}$ \textit{Conv1D(out\_c, out\_c, kernel=5, pad=1)} $\xrightarrow[]{}$ \textit{BatchNorm1D(out\_c)} $\xrightarrow[]{}$ \textit{ReLU()}; \textit{in\_c} denotes the input channel's dimension and \textit{out\_c} denotes the output channel's dimension. Using this block, our model $M$ is defined, as detailed in Table \ref{tab:model_arch}. The output from $\dec^{(i)}$ is then projected to the \textit{number of classes} and followed by a softmax operation to produce probability vectors $\bp^{(i)}$, as described in section 3.2 of the main paper. Our model has a total of $\approx$ 6 million trainable parameters. 

\begin{table}
\begin{center}
\small
\begin{tabular}{|p{0.5cm} | p{1.5cm} | p{3.3cm} | p{1.5cm} |}
\hline
Stage & Input & Model & Output \\\hline

$\enc_0$ & $T_{in} \times 2048$ & \textit{double\_conv(2048, 256)} & $T_{in} \times 256$ \\\hline

\vspace{0.1pt} $\enc_1$ & \vspace{0.1pt} $T_{in} \times 256$ & \vtop{\hbox{\textit{MaxPool1D(2)}}\textit{double\_conv(256, 256)}} & \vspace{0.1pt} $\frac{T_{in}}{2} \times 256$ \\\hline

\vspace{0.1pt} $\enc_2$ & \vspace{0.1pt} $\frac{T_{in}}{2} \times 256$ & \vtop{\hbox{\textit{MaxPool1D(2)}}\textit{double\_conv(256, 256)}} & \vspace{0.1pt} $\frac{T_{in}}{4} \times 256$ \\\hline

\vspace{0.1pt} $\enc_3$ & \vspace{0.1pt} $\frac{T_{in}}{4} \times 256$ & \vtop{\hbox{\textit{MaxPool1D(2)}}\textit{double\_conv(256, 128)}} & \vspace{0.1pt} $\frac{T_{in}}{8} \times 128$ \\\hline

\vspace{0.1pt} $\enc_4$ & \vspace{0.1pt} $\frac{T_{in}}{8} \times 128$ & \vtop{\hbox{\textit{MaxPool1D(2)}}\textit{double\_conv(128, 128)}} & \vspace{0.1pt} $\frac{T_{in}}{16} \times 128$ \\\hline

\vspace{0.1pt} $\enc_5$ & \vspace{0.1pt} $\frac{T_{in}}{16} \times 128$ & \vtop{\hbox{\textit{MaxPool1D(2)}}\textit{double\_conv(128, 128)}} & \vspace{0.1pt} $\frac{T_{in}}{32} \times 128$ \\\hline

\vspace{0.1pt} $\enc_6$ & \vspace{0.1pt} $\frac{T_{in}}{32} \times 128$ & \vtop{\hbox{\textit{MaxPool1D(2)}}\textit{double\_conv(128, 128)}} & \vspace{0.1pt} $\frac{T_{in}}{64} \times 128$ \\\hline

\vspace{2pt} $\bottle$ & \vspace{2pt} $\frac{T_{in}}{64} \times 128$ & \vtop{\hbox{\textit{MaxPool1D(2, 3, 5, 6)}} \textit{conv1d(in\_c=132, out\_c=132, k=3, p=1)}} & \vspace{2pt} $\frac{T_{in}}{64} \times 132$ \\\hline

\vspace{2pt} $\dec_1$ & \vtop{\hbox{$\frac{T_{in}}{64} \times 132$} $\frac{T_{in}}{32} \times 128$} & \vtop{\hbox{\textit{Upsample1D(2)}} \hbox{\textit{concat\_$\enc_5$(132, 128)}} \textit{double\_conv(260, 128)}} & \vspace{2pt} $\frac{T_{in}}{32} \times 128$ \\\hline

\vspace{2pt} $\dec_2$ & \vtop{\hbox{$\frac{T_{in}}{32} \times 128$} $\frac{T_{in}}{16} \times 128$} & \vtop{\hbox{\textit{Upsample1D(2)}} \hbox{\textit{concat\_$\enc_4$(128, 128)}} \textit{double\_conv(256, 128)}} & \vspace{2pt} $\frac{T_{in}}{16} \times 128$ \\\hline

\vspace{2pt} $\dec_3$ & \vtop{\hbox{$\frac{T_{in}}{16} \times 128$} $\frac{T_{in}}{8} \times 128$} & \vtop{\hbox{\textit{Upsample1D(2)}} \hbox{\textit{concat\_$\enc_3$(128, 128)}} \textit{double\_conv(256, 128)}} & \vspace{2pt} $\frac{T_{in}}{8} \times 128$ \\\hline

\vspace{2pt} $\dec_4$ & \vtop{\hbox{$\frac{T_{in}}{8} \times 128$} $\frac{T_{in}}{4} \times 256$} & \vtop{\hbox{\textit{Upsample1D(2)}} \hbox{\textit{concat\_$\enc_2$(128, 256)}} \textit{double\_conv(384, 128)}} & \vspace{2pt} $\frac{T_{in}}{4} \times 128$ \\\hline

\vspace{2pt} $\dec_5$ & \vtop{\hbox{$\frac{T_{in}}{4} \times 128$} $\frac{T_{in}}{2} \times 256$} & \vtop{\hbox{\textit{Upsample1D(2)}} \hbox{\textit{concat\_$\enc_1$(128, 256)}} \textit{double\_conv(384, 128)}} & \vspace{2pt} $\frac{T_{in}}{2} \times 128$ \\\hline

\vspace{2pt} $\dec_6$ & \vtop{\hbox{$\frac{T_{in}}{2} \times 128$} $T_{in} \times 256$} & \vtop{\hbox{\textit{Upsample1D(2)}} \hbox{\textit{concat\_$\enc_0$(128, 256)}} \textit{double\_conv(384, 128)}} & \vspace{2pt} $T_{in} \times 128$ \\\hline
\end{tabular}
\end{center}
\caption{Encoder-Decoder Architecture $M = (\enc, \bottle, \dec)$}
\label{tab:model_arch}
\end{table}

\begin{table}
    \begin{center}
    \small{
    \begin{tabular}{cc|cc|cc}
        \hline
            \multicolumn{2}{c|}{$w_0=5$} & \multicolumn{2}{c|}{$w_0=10$} & \multicolumn{2}{c}{$w_0=20$} \\ \hline
            Edit & MoF & Edit & MoF & Edit & MoF \\
             \hline
            64.7 & 74.6 & 68.9 & 76.6 & 64.3 & 74.1 \\\hline
        \end{tabular} 
    }\end{center}
    \caption{Breakfast variations with Base-Window ($w_0$)}
    \label{tab:bf_base_window}
\end{table}

\begin{table}[]
\begin{center}
\small{
\begin{tabular}{l|c|c|c}
    \hline
    Duration & $\leq\!1$ min & $>\!1$ and $\le\!2.5$ & $>\!2.5$ min \\
    \hline
    No. of Videos & $534$ & $584$ & $594$ \\
    \hline
    MSTCN\cite{li2020ms} & $68.7$ & $70.5$ & $70.2$\\
    \hline
    Ours C2F-TCN & $68.9$ & $69.8$ & $69.7$ \\
    \textbf{(+)} FA-Train & $72.9$ & $72.9$ & $72.7$ \\
    \textbf{(+)} FA-Train-Test & $\mathbf{73.0}$ & $\mathbf{73.3}$ & $\mathbf{75.9}$ \\
    \hline
\end{tabular}
}\end{center}
\caption{MoF for varying lengths of videos in Breakfast.}
\label{tab:length_exp}
\end{table}

\begin{table}[]
    \begin{center}
    \small{
    \begin{tabular}{l|ccccc}
    \hline
    Method & \multicolumn{3}{c}{F1@\{10, 25, 50\}} & Edit & MoF \\
    \hline
         MSTCN & 80.7 & 78.5 & 70.1 & 74.3 & 83.7 \\
         MSTCN\textbf{(+)}Ens. & 73.0 & 71.1 & 64.7 & 66.3 & 83.3 \\
    \hline
    \end{tabular}
    }\end{center}
    \caption{Ensemble of multiple layers of MSTCN~\cite{li2020ms} architecture is not useful as it contains video representations of same temporal resolution without diversity required for ensembling.}
    \label{tab:mstcn_ensemble}
\end{table}

\begin{table}[]
    \begin{center}
    \small{
    \begin{tabular}{l|ccccc}
    \hline
    Method & \multicolumn{3}{c}{F1@\{10, 25, 50\}} & Edit & MoF \\
    \hline
          $\mathcal{L}_{\text{CE}}$ & 83.2 & 80.8 & 71.3 & 73.3 & 84.1 \\
          $\mathcal{L}_{\text{CE}}$ + $\mathcal{L}_{\text{TR}}$ & 84.3 & 81.7 & 72.8 & 76.5 & 84.5\\
         \hline
    \end{tabular}
    }\end{center}
    \caption{Loss function ablation with C2F-TCN on 50salads.}
    \label{tab:loss_function}
\end{table}

\subsection{Training hyper-parameters used}
For all three datasets, Breakfast, 50Salads and GTEA, we use features pre-extracted from an I3D model~\cite{carreira2017quo} pre-trained on Kinetics, and follow the $k$-fold cross-validation averaging to report our final results. Here, $k=\{4, 5, 4\}$ for Breakfast, 50Salads and GTEA, respectively. 
The evaluation metrics and features follow the convention of other recent temporal video segmentation methods~\cite{li2020ms, wang2020boundary}. The feature augmentation's base sampling window $w_0$ is $\{10, 20, 4\}$ for Breakfast, 50Salads and GTEA, respectively, in the supervised and semi-supervised setup. The training hyperparameters for the different datasets and setups are summarized in ~\cref{tab:hyperparameters}.

\begin{table*}
\begin{center}
    \small{
    \begin{tabular}{l|l|cccc|cccc|cccc}
    \hline
         & & \multicolumn{4}{c|}{Breakfast} 
        & \multicolumn{4}{c|}{50Salads} 
        & \multicolumn{4}{c}{GTEA}\\
        \cline{3-14}
        Supervision & Step & LR & WD & Eps. & BS 
                  & LR & WD & Eps. & BS
                  & LR & WD & Eps. & BS\\
        \hline
        Full & & 1e-4 & 3e-3 & 600 & 100 
                     & 3e-4 & 1e-3 & 600 & 25 
                     & 5e-4 & 3e-4 & 600 & 11 \\
        \hline
        Unsupervised & \textit{Contrast step} (model $\mathbf{M}$) & 1e-3 & 3e-3 & 100 & 100 
                     & 1e-3 & 1e-3 & 100 & 50 
                     & 1e-3 & 3e-4 & 100 & 21 \\
        \hline
        \multirow{2}{*}{Semi} & \textit{Classify step} (classifier $\bG$) & 1e-2 & 3e-3 & 700 & 100
                            & 1e-2 & 1e-3 & 1800 & 5 
                            & 1e-2 & 3e-4 & 1800 & 5 \\
        & \textit{Classify step} (model $\mathbf{M}$) & 1e-5 & 3e-3 & 700 & 100
                                  & 1e-5 & 1e-3 & 1800 & 5 
                                  & 1e-5 & 3e-4 & 1800 & 5 \\
    \hline
    \end{tabular}
    }\end{center}
    \caption{The training hyperparameters' learning rate (LR), weight-decay (WD), epochs (Eps.) and batch size (BS) used for the different datasets for full, unsupervised and semi-supervised learning.}\label{tab:hyperparameters}
\end{table*}

\begin{table*}[t!]
    \begin{center}
    \small{
    \begin{tabular}{l|ccccc}
    \hline
         Dataset & F1@10 & F1@25 & F1@50 & Edit & MoF \\
         \hline
         Breakfast & 71.9 $\pm$ 0.6 & 68.8 $\pm$ 0.7 & 58.5 $\pm$ 0.8 & 68.9 $\pm$ 1.3 &  76.6 $\pm$ 0.9 \\
         50Salads & 84.3 $\pm$ 0.7 & 81.7 $\pm$ 0.4 & 72.8 $\pm$ 0.7 & 76.3 $\pm$ 0.8 & 84.5 $\pm$ 0.8 \\
         GTEA & 92.3 $\pm$ 1.1 & 90.1 $\pm$ 0.7 & 80.3 $\pm$ 0.9 & 88.5  $\pm$ 1.5 & 81.2 $\pm$ 0.4 \\
    \hline
  \end{tabular}
  }
\end{center}
    \caption{Mean and standard deviation for our final proposed \textit{C2F-TCN+FA}, reported in Table 6 of the main paper.}\label{tab:error_bars}
    
\end{table*}

\section{Supervised C2F-TCN analysis}\label{sec:ablation_hyperparameter}

\subsection{Choice of the base window $w_0$:} A (too) small $w_0$ leads to a very small range of stochastic windows, \ie{}$[\floor*{\frac{w_0}{2}}, 2w_0]$, and does not allow sufficient training augmentations, while a (too) large $w_0$ completely absorbs (removes) the smaller actions. The ablation results of the base window for the Breakfast dataset are given in Table \ref{tab:bf_base_window}. As the Breakfast dataset has videos with a frame rate of 15 fps, a base window $w_0\!=\!10$ is about 0.67 s. This duration is less than the minimum duration of 99\% of all sub-actions. Similarly, 50Salads has 30 fps features, so we use $w_0=20$. GTEA has many very small segments, sometimes even less than 10 frames (although most are larger than 8 frames), so $w_0=4$ is used for GTEA to capture all actions. 


\subsection{Impact of video length}\label{sec:impact_of_video_length} To obtain a closer look, we split the videos into three length categories and tally the results. To enable a comparison, we train an MSTCN\cite{li2020ms} model, which achieves comparable or higher scores than reported in the original paper for all metrics. ~\cref{tab:length_exp} shows the MoF $\%$ for various video lengths. We observe that after the training augmentation (row 3), the performance improves regardless of the video length. Most notably, for longer videos ($\geq 2.5$ mins), our final proposal with test time augmentation achieves +5.7\% MoF over the MSTCN model.

\subsection{Ablation on loss functions} As discussed in section 4.1 of the main paper, we apply the loss on $\bp^{ens}[t]$ instead of every layer loss (used in~\cite{wang2020boundary, li2020ms}). However, we use the same loss function, cross-entropy loss $\calL_{\text{CE}}$ and transition loss $\mathcal{L}_{\text{TR}}$, as per previous work~\cite{li2020ms, wang2020boundary}. Table ~\cref{tab:loss_function} shows the impact of adding the transition loss on $\bp^{ens}[t]$. The transition loss brings a maximum improvement in the Edit distance scores, which is similar to previous works.

\subsection{Ensemble of the MSTCN layers} 
We verify the improvements from adding C2F ensembling for both our architecture and ED-TCN in Table 2 in the main paper.  

We also try to form an ensemble from the outputs of the different stages of MSTCN~\cite{li2020ms}. However, as shown in ~\cref{tab:mstcn_ensemble}, the ensembling (instead of loss at every layer) curiously 
decreases the original scores. We speculate that there is insufficient diversity in the temporal resolution representations of the MSTCN stages, rendering the ensembled representation less useful. However, as shown in Table 5 of the main paper, feature augmentation does bring improvement in the accuracy and efficiency of the handling sequences for the MSTCN-type architecture.

\subsection{Standard deviations in results} \label{sec:error_bars}~\cref{tab:error_bars} shows the deviations of our final results as reported in Table 6 of the main paper for $4$ runs with different random seeds. For each metric, we report the results in the format $mean \pm std$, \ie the means and the standard deviations for the $4$ runs. For the smallest GTEA dataset, the deviation in the results is higher than in Breakfast and 50Salads.

\subsection{Qualitative examples of segmentation} \label{sec:qualitative_examples} 
\begin{figure*}
    \centering
    \includegraphics[width=\linewidth]{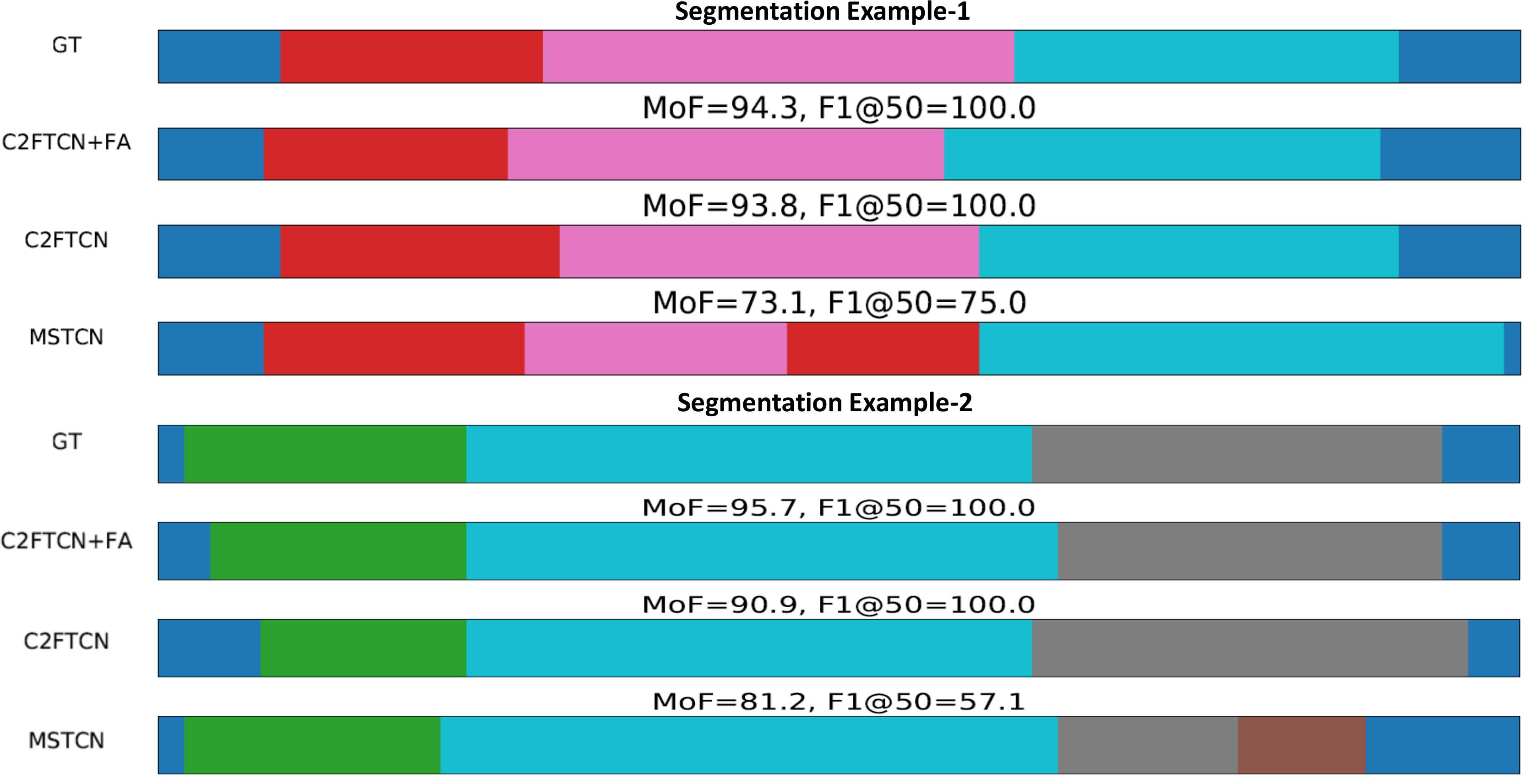}
    \caption{Qualitative examples of segmentation outputs. Different colors represent different actions. We see that our final model C2FTCN+FA best matches the ground truth(GT). MS-TCN gives certain extra segments like ``brown" patch and ``red" patch.}
    \label{fig:qualitative_ex}
\end{figure*}

~\cref{fig:qualitative_ex} visualizes some of the segmentation output (top down) Ground Truth (GT), C2F-TCN with feature augmentation  (\textit{C2FTCN+FA}), from C2FTCN (\ie without augmentation), and finally MSTCN. We compute the corresponding MoF and F1@50 above all the outputs except for GT. We see that C2FTCN, even without feature augmentation, has lower over-segmentation (fragmentation) than MSTCN. Further adding the augmentation \ie{}  (\textit{C2FTCN+FA}) best matches the ground truth segmentation (\textit{GT}).

\section{Unsupervised Representation Analysis}\label{sec:unsupervised_representattion}
\subsection{Multi-resolution features} \label{sec:multi_resolution_feature}

\begin{table*}[h]

\begin{center}
\small{
\begin{tabular}{l|ccccc|ccccc|ccccc}
\hline
& \multicolumn{5}{c|}{Breakfast} 
& \multicolumn{5}{c|}{50Salads} 
& \multicolumn{5}{c}{GTEA}\\
\hline
\textbf{Method} &
\multicolumn{3}{c}{$F1\{10, 25,50\}$} & Edit & MoF & \multicolumn{3}{c}{$F1\{10, 25,50\}$} & Edit & MoF &
\multicolumn{3}{c}{$F1\{10, 25,50\}$} & Edit & MoF \\
\hline

Alternate $\feat^{'}[t]$ &44.3 & 38.3 & 26.1 & 40.9 & 60.9 &
                          32.9 & 27.3 & 19.9 & 26.5 & 51.2 &
                          56.4 & 48.6 & 31.3 & 52.1 & 58.9\\
\textbf{Proposed} $\feat[t]$ & \textbf{57.0} & \textbf{51.7} & \textbf{39.1} & \textbf{51.3} & \textbf{70.5} & 
                     \textbf{40.8} & \textbf{36.2} & \textbf{28.1} & \textbf{32.4} & \textbf{62.5} &
                     \textbf{70.8} & \textbf{65.0} & \textbf{48.0} & \textbf{65.7} & \textbf{69.1} \\
\hline
\end{tabular}
}\end{center}
\caption{Importance of normalization order in the formation of our Multi-Resolution Representation.}\label{tab:multi-res-feature} \label{tab:feature_similarity_formation}
\end{table*}

\subsubsection{Inherent temporal continuity of our feature $\feat$} The inherent temporal continuity encoded in our multi-resolution feature is discussed in Section 5.1.3 of the main paper. For the \textit{nearest} neighbor upsampling strategy, the multi-resolution feature $\feat$ has the property of being similar for nearby frames. Coarser features like $\{\bz_1, \bz_2, \bz_3\}$ are more similar than fine-grained features at higher decoder layers. This also gives independence to the higher resolution features to have high variability even for nearby frames.

Specifically, for two frames $t, s \in \mathbb{N}$, if $\floor*{t/2^u} = \floor*{s/2^u}$ for some integer $u > 0$, then $\simi{\feat[t]}{\feat[s]} \ge 1 - u/3$. This follows from the fact that for nearest upsampling, $\floor*{t/2^u} = \floor*{s/2^u}$ for some $0 \le u \le 5$, implies that
\begin{align}\label{eq.nearest_meaning}
    \bz_v[t] = \bz_v[s] \quad\text{for all}\quad 1\le v \le 6-u.
\end{align}
Meaning, the lower-resolution features coincide with proximal frames. As discussed and shown in equation (15) of the main paper, all the layers make an equal contribution while calculating the similarity of our multi-resolution feature. For $t,s$, with $\floor*{t/2^u} = \floor*{s/2^u}$ for some $0 \le u \le 5$, we start with the equation (15) of the main text to derive --

\begin{align*}
    & \simi{\feat[t]}{\feat[s]}\\ 
    &= \sum_{v=1}^6 \frac{1}{6} \cdot \simi{\bz_v[t]}{\bz_v[s]}\\
    &= \sum_{v=1}^{6-u} \frac{1}{6} \cdot \simi{\bz_v[t]}{\bz_v[s]} + \sum_{v=7-u}^{6} \frac{1}{6} \cdot \simi{\bz_v[t]}{\bz_v[s]}\\
    &= \frac{6-u}{6} + \sum_{v=7-u}^{6} \frac{1}{6} \cdot \simi{\bz_u[t]}{\bz_u[s]} \quad \BK{\text{from \eqref{eq.nearest_meaning}}}\\
    &\ge \frac{6-u}{6} - \frac{u}{6} \qquad\qquad \BK{\text{as}\,\, \simi{\cdot}{\cdot} \ge -1}\\
    &= 1 - \frac{u}{3}.
\end{align*}
That means for $\floor*{t/2^u} = \floor*{s/2^u}$ for some $0 \le u \le 5$ implies $\simi{\feat[t]}{\feat[s]} \ge 1 - \frac{u}{3}$. The inequality is trivial for $u > 5$.

\subsubsection{Normalization:} Our proposed multi-resolution feature, as outlined in Section 5.1.3 of the main paper, is defined for frame $t$ as $\feat[t] = \BK{\bar{\bz}_1[t]:\bar{\bz}_2[t]:\ldots:\bar{\bz}_6[t]}$, where $\bar{\bz}_u[t] = {\hat{\bz}}_u[t]/\norm{{\hat{\bz}}_u[t]}$, \ie ${\hat{\bz}}_u[t]$, the upsampled feature from decoder $u$ is normalized first for each frame and then concatenated along the latent dimension. 
An alternative and naive construction would be to apply normalization after concatenation,~\ie{} $\feat^{'}[t] = \BK{\hat{\bz}_1[t]:\hat{\bz}_2[t]:\ldots:\hat{\bz}_6[t]}$. The features $\hat{\bz}_u$ are the upsampled \textit{un-normalized} feature vector of decoder layer $u$. Note that a final normalization of $\feat^{'}[t]$ is no longer necessary as the cosine similarity is invariant.  We verify in \ref{tab:feature_similarity_formation} that applying normalization \textit{before} concatenation is critical.

\begin{table*}[t]

\begin{center}
\small{
\begin{tabular}{l|ccccc|ccccc|ccccc}
\hline
& \multicolumn{5}{c|}{Breakfast} 
& \multicolumn{5}{c|}{50Salads} 
& \multicolumn{5}{c}{GTEA}\\
\hline
\textbf{Method} &
\multicolumn{3}{c}{$F1\{10, 25,50\}$} & Edit & MoF & \multicolumn{3}{c}{$F1\{10, 25,50\}$} & Edit & MoF &
\multicolumn{3}{c}{$F1\{10, 25,50\}$} & Edit & MoF \\
\hline
No Augment & 55.6 & 50.2 & 36.5 & 49.4 & 69.4 & 
              40.0 & 34.1 & 27.0 & 31.0 & 62.3 &
              70.0 & 63.4 & 47.2 & 65.6	& 69.0 \\
Augment & \textbf{57.0} & \textbf{51.7} & \textbf{39.1} & \textbf{51.3} & \textbf{70.5} & 
 \textbf{40.8} & \textbf{36.2} & \textbf{28.1} & \textbf{32.4} & \textbf{62.5} &
 \textbf{70.8} & \textbf{65.0} & \textbf{48.0} & \textbf{65.7} & \textbf{69.1} \\
 

\hline
\end{tabular}
}\end{center}
\caption{Impact of using Temporal Feature Augmentation Strategy on Unsupervised Representation Learning} \label{tab:feature-agument}
\end{table*}
\begin{table}[t]
    \begin{center}
    \small{
    
    \begin{tabular}{l|ccccc}
    \hline
         Samples($2K$) & \multicolumn{3}{c}{$F1\{10, 25,50\}$} & Edit & MoF \\\hline
         60 & 38.5 & 33.2 & 25.1 & 29.9 & 62.5\\
         120 & 40.8	& 36.2 & 28.1 & 32.4 & 62.5\\
         180 & 39.1	& 34.5 & 27.3 & 30.8 & 62.1\\\hline
    \end{tabular}
    }\end{center}
    \caption{Ablation results of the number of samples per video required for representation learning.}
    \label{tab:num_samples}
\end{table}

\begin{table}[]
    \begin{center}
    \small{
    \begin{tabular}{cccc}
        \hline
         Type (Number) & Breakfast & 50Salads & GTEA \\
         \hline
         FINCH ($A$)  & 61.7 & 56.3 & 56.7 \\
         Kmeans ($A$) & 70.0 & 60.4 & 65.6 \\
         \textbf{Kmeans ($\approx 2A$)} & \textbf{70.5} & \textbf{62.5} & \textbf{69.1} \\
        \hline
    \end{tabular}
    }\end{center}
    \caption{Unsupervised Representation's MoF variation with different clustering types and number of clusters used during training. $A$ denotes the number of unique actions in the dataset.}
    \label{tab:kmeans_finch}
\end{table}

\subsection{Sampling strategy, number of samples $2K$} We show ablations for the choice of $2K$ (\ie number of representation samples drawn per video, as described in section 5.1.1 of main paper) in ~\cref{tab:num_samples}. Thus, our value of $2K$ is determined by experimental validation. In 50salads with $2K=120$ and a batch size of 50, we obtain roughly 0.6 million positive samples per batch, with each positive sample having roughly around 6.5K negative samples. 

\subsection{Impact of Temporal Feature Augmentation} In Table~\ref{tab:feature-agument} we show the improvements in unsupervised features linear evaluation scores when training with feature-augmentation(FA). 

\begin{table}
    \begin{center}
    \small{
    \begin{tabular}{l|ccccc}
    \hline
         \textbf{Method} & \multicolumn{3}{c}{$F1@\{10,25,50\}$} & Edit & MoF\\\hline
         Supervised & 30.5 & 25.4 & 17.3 & 26.3 & 43.1 \\
         ICC-wo-unsupervised & 42.6	& 37.5 & 25.3 & 35.2 & 53.4 \\
         ICC-with-unsupervised & \textbf{52.9} & \textbf{49.0} & \textbf{36.6} & \textbf{45.6} & \textbf{61.3} \\
    \hline
    \end{tabular}
    }\end{center}
    \caption{``ICC-wo-unsupervised'' (removing the initial unsupervised representation learning from ICC) on 50Salads with 5\% $\calD_L$. The ICC results are from the fourth iteration \ie ($\text{ICC}_4$).}
    \label{tab:wo_unsuper_pretrain}
\end{table}

\begin{table}
\begin{center}
\small{
\begin{tabular}{l|ccccc}
    \hline
     & F1@10 & F1@25 & F1@50 & Edit & MoF \\
    \hline
    Unsupervised & 40.8 & 36.2 & 28.1 & 32.4 & 62.5 \\
    $\text{ICC}_\text{2}$ & 51.3 & 46.6 & 36.5 & 44.7 & 61.3 \\
    $\text{ICC}_\text{3}$ & 52.5 & 47.2 & 36.5 & 45.4 & 62.1 \\
    $\text{ICC}_\text{4}$ & 52.6 & 47.7 & 38.1 & 46.7 & 61.3 \\
    \hline
    \end{tabular}
}\end{center}
\caption{Improvement in representation on 50Salads for 5\% labelled data with more iterations of ICC. Note: Representation is evaluated with 100\% data with a simple Linear Classifier, as discussed in section 4.4.}
\label{tab:50Salads_reprentation_learning}
\end{table}

\subsection{Input feature clustering} 
As described in section 5.1.1 of the main paper, unsupervised feature learning requires cluster labels from the input features. 
We cluster at the mini-batch level with a standard {$k$-means} and then compare with 
Finch~\cite{sarfraz2019efficient}, an agglomerative clustering that has been shown to be useful in unsupervised temporal segmentation \cite{sarfraz2021temporally}. 
Comparing the two in ~\cref{tab:kmeans_finch}, we observe that 
$K$-means performs better.  We speculate that this is because 
Finch is designed for 
per-video clustering. In contrast, our clustering on the mini-batch is on a dataset level, i.e., over multiple video sequences of different complex activities. 

To choose $k$ in the $k$-means clustering, we choose $\approx 2C$ ($C$ denotes the number of unique actions) number of clusters, resulting in $K=\{100, 40, 30\}$ for the Breakfast, 50Salads, and GTEA datasets, respectively.  The advantage of using  $\approx 2C$ clusters versus simply $C$ is verified in ~\cref{tab:kmeans_finch}. The improvement is greater for datasets with fewer action classes like GTEA and 50Salads than the Breakfast action dataset.

\section{Semi-Supervised Learning Analysis}\label{sec:semi_supervised}

\begin{table*}
\begin{center}
\small{
\begin{tabular}{c|c|ccccc|ccccc|ccccc}
\hline
& & \multicolumn{5}{c|}{Breakfast} & \multicolumn{5}{c|}{50Salads} & \multicolumn{5}{c}{GTEA} \\
\hline
\%$D_L$ &\textbf{Method} & \multicolumn{3}{c}{$F1@\{10,25,50\}$} & Edit & MoF & \multicolumn{3}{c}{$F1@\{10,25,50\}$} & Edit & MoF & \multicolumn{3}{c}{$F1@\{10,25,50\}$} & Edit & MoF \\
\hline
\multirow{4}{*}{\textbf{$\approx$10}} & $\text{ICC}_1$ 
                    & 57.0 & 51.9 & 36.3 & 56.3 & 65.7 
                    & 51.1 & 45.6 & 34.5 & 42.8 & 65.3 
                    & 82.2 & 78.9 & 63.8 & 75.6 & 72.2\\
& $\text{ICC}_2$  
                    & 60.0 & 54.5 & 38.8 & 59.5 & 66.7
                    & 56.5 & 51.6 & 39.2 & 48.9 & 67.1 
                    & 83.4 & 80.1 & 64.2 & 75.9 & 72.9 \\
& $\text{ICC}_3$  
                    & 62.3 & 56.5 & 40.4 & 60.6 & 67.8 
                    & 60.7 & 56.9 & 45.0 & 52.4 & 68.2
                    & 83.5 & 80.8 & 64.5 & 76.3 & 73.1\\
& $\text{ICC}_4$ 
                    & 64.6 & 59.0 & 42.2 & 61.9 & 68.8
                    & 67.3 & 64.9 & 49.2 & 56.9 & 68.6 
                    & 83.7 & 81.9 & 66.6 & 76.4 & 73.3 \\
                    
\cline{2-17}
& Gain & 7.6 & 7.1 & 5.9 & 5.6 & 3.1 
       & 16.2 & 19.3 & 14.7 & 14.1 & 3.3 
       & 1.5 & 3.0 & 2.8 & 0.8 & 1.1 \\
\hline

\end{tabular}
}\end{center}
\caption{Quantitative evaluation of progressive semi-supervised improvement with more iterations of ICC with $\approx$ 10\% labelled training videos.}
\label{tab:icc_10_100}
\end{table*}

\begin{table*}

\begin{center}
\small{
\begin{tabular}{l|l|ccccc}
    \hline
     Dataset & ICC(Num Videos) & F1@10 & F1@25 & F1@50 & Edit & MoF \\
    \hline
    \multirow{4}{*}{Breakfast}  & $\text{ICC}_1$ ($\approx 63$ Videos) & 54.5 $\pm$  1.2 & 48.7 $\pm$ 1.1  & 33.3 $\pm$ 1.1 & 54.6 $\pm$ 0.9 & 64.2 $\pm$  1.3 \\
    & $\text{ICC}_4$ ($\approx 63$ videos) &  60.2 $\pm$ 1.5 & 53.5 $\pm$ 1.3 & 35.6 $\pm$ 0.9 & 56.6 $\pm$ 1.2 & 65.3 $\pm$ 1.8 \\
    \cline{2-7}
    & $\text{ICC}_1$ ($\approx 120$ Videos) & 57.0 $\pm$ 1.9 & 51.9 $\pm$  2.1 & 36.3 $\pm$ 1.3 & 56.3 $\pm$ 1.2 & 65.7 $\pm$ 1.9 \\
    & $\text{ICC}_4$ ($\approx 120$ Videos) &  64.6 $\pm$ 2.1 & 59.0 $\pm$ 1.9 & 42.2 $\pm$ 2.5 & 61.9 $\pm$ 2.2 & 68.8 $\pm$ 1.3 \\\hline

    \multirow{4}{*}{50salads} &  $\text{ICC}_1$ (3 Videos) & 41.3 $\pm$ 1.9 & 37.2 $\pm$ 1.5 & 27.8 $\pm$ 1.1 & 35.4 $\pm$ 1.6 & 57.3 $\pm$ 2.3 \\
    &  $\text{ICC}_4$ (3 videos) &  52.9 $\pm$ 2.2 & 49.0 $\pm$ 2.2 & 36.6 $\pm$ 2.0 & 45.6 $\pm$ 1.4 & 61.3 $\pm$ 2.3 \\
    \cline{2-7}
    &  $\text{ICC}_1$ (5 Videos) & 51.1 $\pm$  2.1 & 45.6 $\pm$  1.3 & 34.5 $\pm$ 1.7 & 42.8 $\pm$ 1.1 & 65.3 $\pm$  0.8 \\
    & $\text{ICC}_4$  (5 videos) & 67.3 $\pm$ 1.8 & 64.9 $\pm$ 2.5 & 49.2 $\pm$ 1.8 & 56.9 $\pm$ 2.1 & 68.6 $\pm$  0.7 \\
    \hline
    
    \end{tabular}
}\end{center}
\caption{Mean and standard deviation for 5 different selections of 5\% and 10\% labelled videos from Breakfast and 50Salads. For each metric we report the results in the format $mean \pm std$, \ie the means and the standard deviation for the 5 runs.}
\label{tab:50Salads_error_bars_5}
\end{table*}
\subsection{ICC without unsupervised step}\label{subsec:icc_wo_pretraining}

In ~\cref{tab:wo_unsuper_pretrain}, we show the results of our ICC without the initial \textit{``unsupervised representation learning''} as \textbf{``ICC-Wo-Unsupervised''}. This essentially means that the $2^{nd}$ row of ~\cref{tab:wo_unsuper_pretrain} represents the scenario in which we remove from our ICC algorithm the $1^{st}$ contrast step that is learned with cluster labels. The improvement in scores over the supervised setup is quite low compared to the full-ICC \textit{with} the unsupervised pre-training, shown as \textbf{``ICC-With-Unsupervised''}. This verifies the importance of our unsupervised learning step in ICC.

\subsection{Iterative progression of ICC results} \label{subsec:icc_progression_results}

We discuss our detailed semi-supervised algorithm in section 5.2 of our main paper and provide a visualization of the algorithm in Figure 7. 

\begin{figure*}
    \centering
    \includegraphics[width=0.9\linewidth]{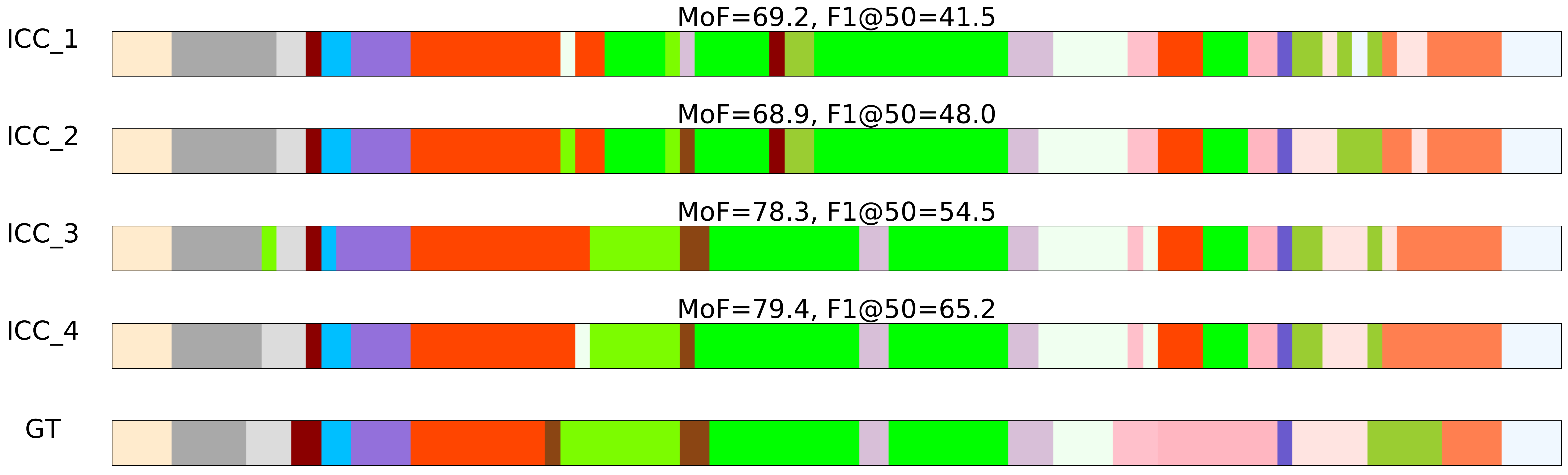}
    \caption{A qualitative example taken from  50Salads, showing progressive improvement in segmentation results with number of iterations of ICC. Some segments become more aligned to ground truth (GT), leading to improved MoF and F1@50 scores.}
    \label{fig:50salads_icc_example}
\end{figure*}

\subsubsection{Improvement after the \textit{Contrast} step}
In Table \ref{tab:50Salads_reprentation_learning} we show the improvement in representation after each \textit{contrast} step. Due to the usage of better pseudo-labels obtained from the preceding \textit{classify} step, the following contrast step results in better representations as more iterations are performed. For the 5\% labelled videos of the 50Salads dataset, we can see that there is a clear improvement in the F1 and Edit scores as more iterations are performed. Note that the evaluation of the learned representation is linear evaluation protocol as described in section 5.1.4 of the main text.

\subsubsection{Improvement after the \textit{Classify} step}
In Table 12 of the main text, we show the progressive improvement in performance for the 5\% labelled videos, evaluated after the \textit{classify} step of each ICC iteration. In ~\cref{tab:icc_10_100}, we show the same progressive improvements for the 10\% videos. The evaluation is done after the \textit{classify} step of each iteration of the algorithm. Our ICC raises the overall scores on all datasets, with stronger improvements in the F1 and Edit scores. 

\subsection{Qualitative visualization of segmentation} In ~\cref{fig:50salads_icc_example} we use an example of the segmentation results from the 50Salads dataset to show how the segmentation results improve (become more aligned with GT with increase in MoF and F1@50) with more iterations of ICC.   
\subsection{Standard deviations in results} \label{subsec:mean_std_results} We show our standard deviations in results for the 50Salads and Breakfast datasets for variations in labelled data used in ~\cref{tab:50Salads_error_bars_5}. We show the variation in results for $\text{ICC}_1$ and $\text{ICC}_4$ when we take 5 different random selections of 5\%, 10\% labelled videos in Breakfast and 50Salads from the corresponding training splits. We report the means and standard deviations for the different choices in the $mean \pm std$ format.

\section*{Acknowledgements} The authors would like to thank
National Research Foundation, Singapore (under its AI Singapore Programme (AISG Award No: AISG2-RP-2020-016)) for the research support. Any opinions, findings and conclusions or recommendations expressed in this material are those of the author(s) and do not reflect the views of National Research Foundation, Singapore.


\ifCLASSOPTIONcaptionsoff
  \newpage
\fi



\bibliographystyle{IEEEtran}
\bibliography{IEEEabrv,main}
%



%
\vfill
\begin{IEEEbiography}[{\includegraphics[width=1in,height=1in,clip,keepaspectratio]{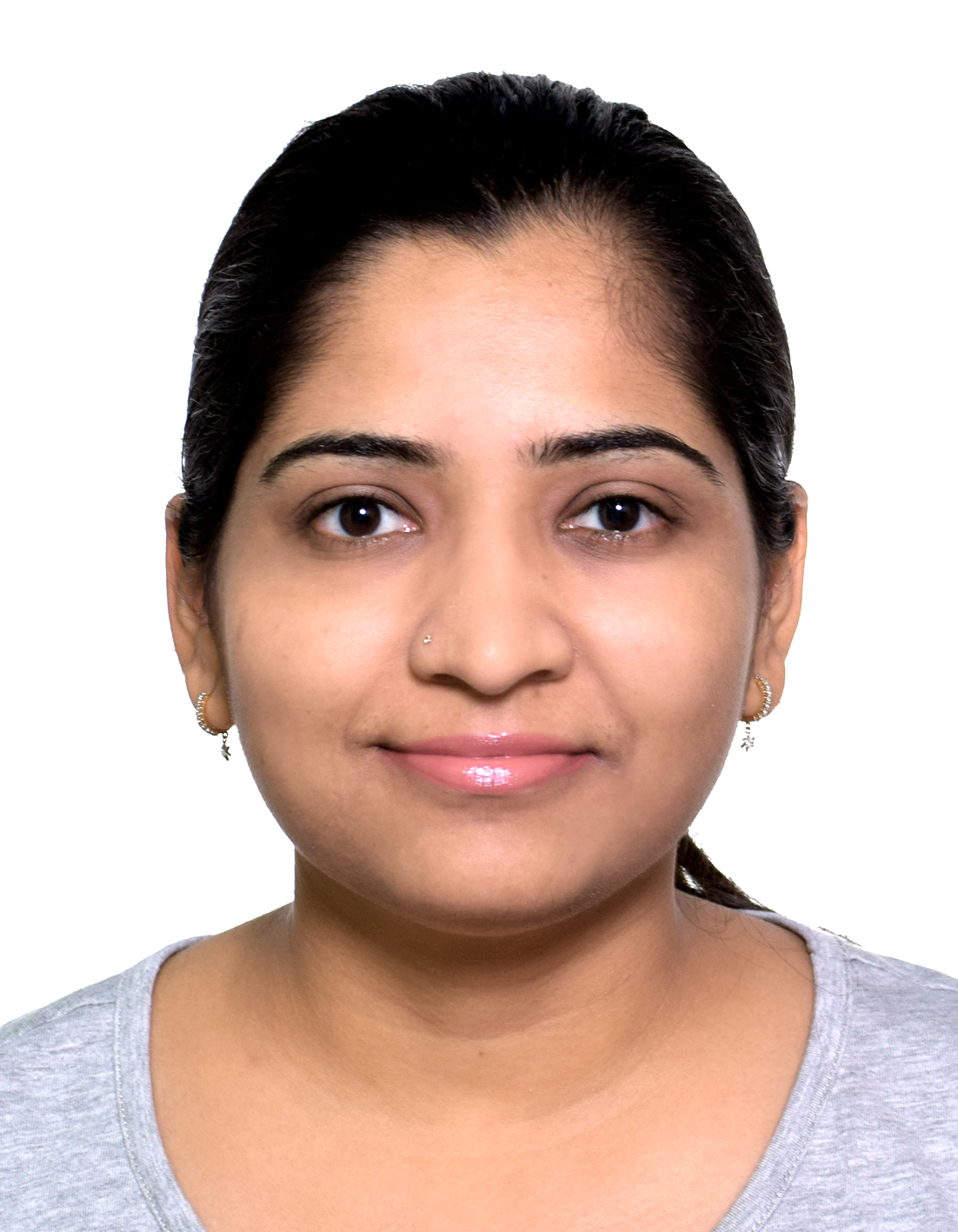}}]{Dipika Singhania}
is a PhD candidate at School of Computing, National University of Singapore. She received
his bachelor degree from IIEST, India in
2013. Her research interests include video recognition and segmentation using deep learning.

\end{IEEEbiography}
\vfill
\begin{IEEEbiography}[{\includegraphics[width=1in,height=1in,clip,keepaspectratio]{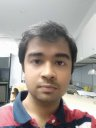}}]{Rahul Rahaman} is a PhD candidate at Department of Statistics and Data Science, National University of Singapore. He received
his masters and bachelor degree from Indian Statistical Institute in
2011 and 2013 respectively. His research interests include uncertainty quantification and use of less supervision in videos and images.
\end{IEEEbiography}

\vfill
\begin{IEEEbiography}[{\includegraphics[width=1in,height=1in,clip,keepaspectratio]{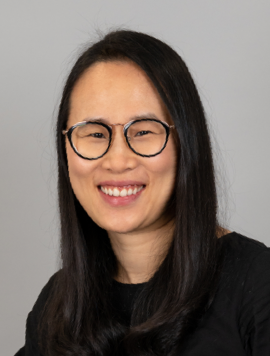}}]{Angela Yao} is a Assistant Professor in Computer Science at the School of Computing since 2018, where she leads the Computer Vision and Machine Learning group. She works on topics ranging from segmentation, pose estimation, to video understanding. Before NUS, she was a junior professor at the University of Bonn, Germany. She received her PhD in 2012 from ETH Zurich.
\end{IEEEbiography}




\end{document}